\documentclass[final]{cvpr}

\usepackage{times}
\usepackage{epsfig}
\usepackage{graphicx}
\usepackage{amsmath}
\usepackage{amssymb}

\usepackage{booktabs}
\usepackage{color}
\usepackage{xspace}
\usepackage{cite}  % orders citations based on number: [16,1,2] -> [1,2,16]
\usepackage{array}
\usepackage{rotating}
\newcolumntype{M}[1]{>{\centering\arraybackslash}m{#1}}
\usepackage{tabulary}
\usepackage{array, multirow}
\usepackage{xcolor}

\usepackage[pagebackref=true,breaklinks=true,colorlinks,bookmarks=false]{hyperref}

\newcommand{\blue}[1]{\textcolor{blue}{#1}}
\newcommand{\PAR}[1]{\vskip4pt \noindent{\bf #1~}}

\newcommand{\thickline}{$\mathrel{\vcenter{\hbox{\rule{15pt}{2pt}}}}$ }
\newcommand{\dashine}{$\mathrel{\vcenter{\hbox{\rule{4pt}{2pt}}}}$
$\mathrel{\vcenter{\hbox{\rule{4pt}{2pt}}}}$ $\mathrel{\vcenter{\hbox{\rule{4pt}{2pt}}}}$}

\newcommand{\cstick}[1]{{\color{#1} \thickline}}
\newcommand{\cdash}[1]{{\color{#1} \dashine}}

\definecolor{cgreen}{RGB}{26, 110, 53}
\definecolor{cgrass}{RGB}{123, 252, 3}
\definecolor{cbrown}{RGB}{161, 100, 56}
\definecolor{cyellow}{RGB}{237, 187, 36}
\definecolor{cpurple}{RGB}{177, 87, 250}
\definecolor{cpurblue}{RGB}{194, 207, 242}
\definecolor{cgrey}{RGB}{157, 163, 163}
\definecolor{corange}{RGB}{245, 130, 69}
\definecolor{cblue}{RGB}{66, 120, 245}
\definecolor{csky}{RGB}{148, 250, 255}
\definecolor{ccyan}{RGB}{8, 189, 171}
\definecolor{crose}{RGB}{235, 101, 157}
\definecolor{cpink}{RGB}{255, 212, 212}
\definecolor{cred}{RGB}{219, 15, 15}
\definecolor{cdark}{RGB}{0, 0, 0}

\newcommand{\hgreen}[1]{\textcolor{cgreen}{#1}\xspace}
\newcommand{\hcyan}[1]{\textcolor{ccyan}{#1}\xspace}
\newcommand{\hblue}[1]{\textcolor{cblue}{#1}\xspace}

\def\cvprPaperID{2491} % *** Enter the CVPR Paper ID here
\def\confYear{CVPR 2021}

\begin{document}

%%%%%%%%% TITLE
\title{Patch2Pix: Epipolar-Guided Pixel-Level Correspondences}
\author{Qunjie Zhou$^1$ \,\, Torsten Sattler$^2$ \,\, Laura Leal-Taix\'{e}$^1$ 
\and 
$^1$Technical University of Munich \,\,\,
$^2$CIIRC, Czech Technical University in Prague
\thanks{This research was funded by the
Humboldt Foundation through the Sofja Kovalevskaya Award, 
the EU Horizon 2020 project RICAIP (grant agreeement No. 857306), 
and the European Regional Development Fund under project IMPACT (No.~CZ.02.1.01/0.0/0.0/15 003/0000468).}
}

\maketitle
%\thispagestyle{empty}
%\pagestyle{empty}

%%%%%%%% ABSTRACT
\begin{abstract}
%Deep learning has been applied to \qj{the} 
The classical matching pipeline used for visual localization typically involves three steps: (i) local feature detection and description, (ii) feature matching, and (iii) outlier rejection. 
Recently emerged correspondence networks propose to perform those steps inside a single network but suffer from low matching resolution due to the memory bottleneck. 
In this work, we propose a new perspective to estimate correspondences in a \textit{detect-to-refine} manner, where we first predict patch-level match proposals and then refine them. 
We present {\it Patch2Pix}, a novel refinement network that refines match proposals by regressing pixel-level matches from the local regions defined by those proposals and jointly rejecting outlier matches with confidence scores. Patch2Pix is weakly supervised to learn correspondences that are consistent with the epipolar geometry of an input image pair. 
We show that our refinement network significantly improves the performance of correspondence networks on image matching, homography estimation, and localization tasks.
In addition, we show that our learned refinement generalizes to fully-supervised methods without re-training, which leads us to state-of-the-art localization performance.
The code is available at \url{https://github.com/GrumpyZhou/patch2pix}.

\end{abstract}

%%%%%%%%% BODY TEXT

\section{Introduction}
%%%%%%%%%Teaser%%%%%%%%%%
 \begin{figure}[t]
 \centering
   \includegraphics[width=0.48\textwidth]{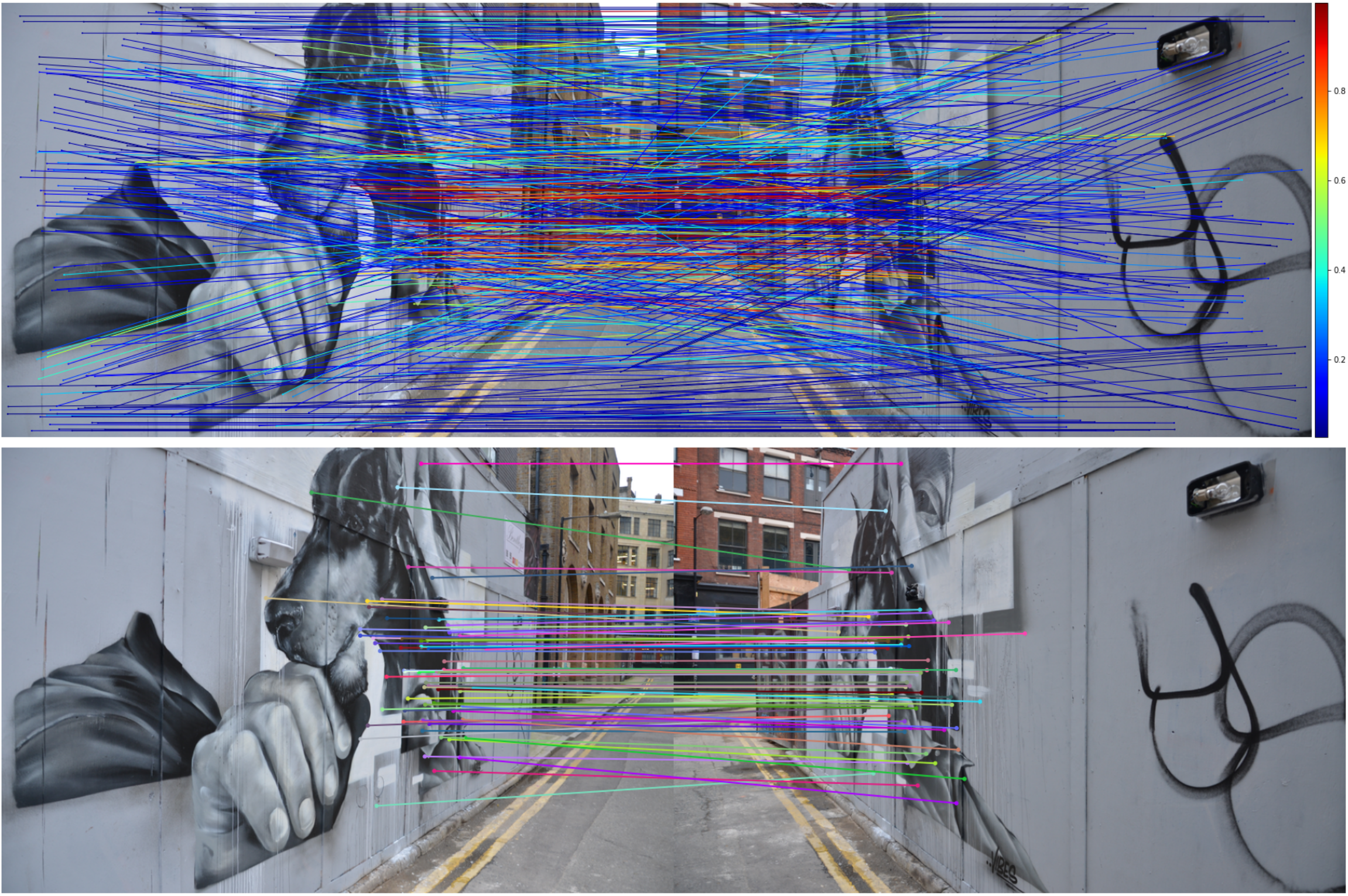}
   \caption{\textbf{An example of {\it Patch2Pix} correspondences.} 
   %In this paper, we propose a refinement network that improves the quality of match proposals obtained from another method by simultaneously rejecting the outliers and relocate pixel-level accurate matches. 
   In the top figure, the matches refined by {\it Patch2Pix} are coloured according to the predicted confidence scores. The less confident matches (in \blue{blue}) appear mostly on the road or the blank wall. In the bottom figure, we show that the inlier matches can well handle the large viewpoint change. We show more quantitative results for handling various challenging conditions in the supp. mat (\cf Sec.\ref{sec:supp:qualitative}). }
   \label{fig:teaser}
 \end{figure}

Finding image correspondences is a fundamental step in several computer vision tasks such as Structure-from-Motion (SfM) \cite{Schoenberger2016CVPR,Wu20133DV} and Simultaneous Localization and Mapping (SLAM)~\cite{MurAtal2015TRO,Engel2018PAMI}.
Given a pair of images, pixel-level correspondences are commonly established through a local feature matching pipeline, which involves the following three steps: i) detecting and describing local features, ii) matching the nearest neighbors using the feature descriptors, and iii) rejecting outlier matches. 

Traditional hand-crafted local features such as SIFT~\cite{Lowe2004IJCV} or SURF~\cite{Bay2008CVIU} are vulnerable to extreme illumination changes, motion blur and repetitive and weakly textured scenes.
Therefore, recent works~\cite{Wang2020ECCV, Luo2020CVPR, Revaud2019NIPS, Ebel2019CVPR, Luo2019CVPR, Detone2018CVPRW, Dusmanu2019CVPR} propose to learn to detect and describe local features using neural networks, showing that learned features can be robustly matched under challenging conditions~\cite{Wang2020ECCV, Luo2020CVPR, Revaud2019NIPS, Dusmanu2019CVPR}.
Instead of focusing on improving local features, ~\cite{Moo2018CVPR, Zhang2019ICCV, Sun2020CVPR, Brachmann2019ICCV} suggest to learn a filtering function from sets of correspondences to reject outlier matches.
A recent method~\cite{Sarlin2020CVPR} further proposes to jointly learn the matching function and outlier rejection via graph neural networks and the Sinkhorn algorithm ~\cite{Sinkhorn1967PJM, Cuturi2013NIPS}.
Combining a learned feature~\cite{Detone2018CVPRW} and learned matcher~\cite{Sarlin2020CVPR} has set the state-of-the-art results on several geometry tasks, showing a promising direction towards a full learnable matching pipeline.

Learning the whole matching pipeline has already been investigated in several works~\cite{Rocco2018NIPS, Rocco2020ECCV, Li20NIPS}, where a single network directly outputs correspondences from an input image pair.
The main challenge faced with those correspondence networks is how to efficiently perform matching while reaching pixel-level accuracy.
In order to keep computation speed and memory footprint manageable, ~\cite{Rocco2017CVPR} has to match at a rather low resolution, which is shown to be less accurate in relative pose estimation~\cite{Zhou2020ICRA}. 
While sparse convolutions have been applied in~\cite{Rocco2020ECCV} to match at higher resolution, they still do not achieve pixel-level matching.
One advantage of the correspondences networks~\cite{Rocco2018NIPS, Rocco2020ECCV} is that they are weakly supervised to maximize the average matching score for a matching pair and minimize it for a non-matching pair, however, they learn less effectively in pixel-level matching.
This is in contrast to methods that require full supervision from ground truth (GT) correspondences~\cite{Luo2020CVPR, Dusmanu2019CVPR, Sarlin2020CVPR, Germain2020ECCV,Revaud2019NIPS, Detone2018CVPRW}.
While the GT correspondences provide very precise signals for training, they might also add bias to the learning process. 
For example, using the sparse keypoints generated by an  SfM pipeline with a specific detector as supervision, 
a keypoint detector might simply learn to replicate these detections rather than learning more general features~\cite{Ono2018NIPS}.
% the network is encouraged to learn the type of features predicted by that specific detector, which prevents the network to learn more general features~\cite{Ono2018NIPS}. 
% \qj{(To replace the last sentence only: For example, the network is encouraged to predict keypoints on the sparse keypoint locations that provied by the SfM, which prevents the network to learn more general features~\cite{Ono2018NIPS}. }
To avoid such type of bias in the supervision, a recent work~\cite{Wang2020ECCV} proposes to use relative camera poses as weak supervision to learn local feature descriptors.
Compared to the mean matching score loss used in~\cite{Rocco2018NIPS, Rocco2020ECCV}, they are more precise by containing the geometrical relations between the images pairs.

%Ours 
In this paper, we propose {\it Patch2Pix}, a new view for the design of correspondence networks.
Inspired by the successful \textit{detect-to-refine} practice in the object detection community~\cite{Ren2015NIPS}, our network first obtains patch-level match proposals and then refines them to pixel-level matches.
See an example of our matches in Fig.~\ref{fig:teaser}. 
%Fig.~\ref{fig:teaser} shows an example of our matches that handles large viewpoint variation.
Our novel refinement network is weakly supervised by epipolar geometry computed from relative camera poses, which are used to regress geometrically consistent pixel-wise matches within the patch proposal.
Compared to ~\cite{Wang2020ECCV}, we optimize directly on match locations to learn matching, while they optimize through matching scores to learn feature descriptors.
Our method is extensively evaluated on a set of geometry tasks, showing state-of-the-art results.
We summarize our {\bf contributions} as: 
i) We present a novel view for finding correspondences, where we first obtain patch-level match proposals and then refine them to pixel-level matches.
ii) We develop a novel match refinement network that jointly refines the matches via regression and rejects outlier proposals. 
It is trained without the need for pixel-wise GT correspondences.
iii) We show that our model consistently improves match accuracy of correspondence networks for image matching, homography estimation and visual localization.
iv) Our model generalizes to fully supervised methods without the need for retraining, and achieves state-of-the-art results on indoor and outdoor long-term localization. %experiments superglue + ours

\section{Related Work}
Researchers have recently opted for leveraging deep learning to detect robust and discriminative local features ~\cite{Detone2018CVPRW, Dusmanu2019CVPR , Revaud2019NIPS, Ebel2019CVPR,  Luo2020CVPR, Wang2020ECCV}.
D2Net~\cite{Dusmanu2019CVPR} detects keypoints by finding local maxima on CNN features at a 4-times lower resolution \wrt the input images, resulting in less accurate detections.
Based on D2Net, ASLFeat~\cite{Luo2020CVPR} uses deformable convolutional networks and extracts feature maps at multiple levels to obtain pixel-level matches.
R2D2~\cite{Revaud2019NIPS} uses dilated convolutions to preserve image resolution and predicts per-pixel keypoints and descriptors, which gains accuracy at the cost of computation and memory usage.
Given the keypoints, CAPS~\cite{Wang2020ECCV} fuses features at several resolutions and obtains per-pixel descriptors by interpolation.
The above methods are designed to learn local features and require a further matching step to predict the correspondences.

\PAR{Matching and Outlier Rejection.}
% ACNet~\cite{Sun2020CVPR}, OANet~\cite{Zhang2019ICCV}, NG-RANSAC~\cite{Brachmann2019ICCV}
% SuperGlue~\cite{Sarlin2020CVPR}, S2DNet~\cite{Germain2020ECCV}
Once local features are detected and described, correspondences can be obtained using Nearest Neighbor (NN) search~\cite{Muja2014PAMI} based on the Euclidean distance between the two feature representations. 
Outliers are normally filtered based on mutual consistency or matching scores.
From a set of correspondences obtained by NN search, recent works~\cite{Moo2018CVPR, Zhang2019ICCV, Sun2020CVPR, Brachmann2019ICCV} learn networks to predict binary labels to identify outliers~\cite{Moo2018CVPR, Zhang2019ICCV, Sun2020CVPR}, or probabilities that can be used by RANSAC~\cite{Fischler81CACM} to weight the input matches~\cite{Brachmann2019ICCV}.
Notice, those methods do not learn the local features for matching and the matching function itself, thus they can only improve within the given set of correspondences.
Recent works further propose to learn the whole matching function~\cite{Sarlin2020CVPR, Germain2020ECCV}.
SuperGlue~\cite{Sarlin2020CVPR} learns to improve SuperPoint~\cite{Detone2018CVPRW} descriptors for matching using a graph neural network with attention and computes the correspondences using the Sinkhorn algorithm ~\cite{Sinkhorn1967PJM, Cuturi2013NIPS}.
S2DNet~\cite{Germain2020ECCV} extracts sparse features at SuperPoint keypoint locations for one image and matches them exhaustively to the dense features extracted for the other image to compute correspondences based on the peakness of similarity scores.
While those methods optimize feature descriptors at keypoint locations specifically for the matching process, they do not solve the keypoint detection problem. 

\PAR{End-to-End Matching.}
Instead of solving feature detection, feature matching, and outlier rejection separately, recently correspondences networks~\cite{Rocco2018NIPS, Rocco2020ECCV, Li20NIPS} have emerged to accomplish all steps inside a single forward pass.
NCNet uses a correlation layer~\cite{Rocco2017CVPR} to perform the matching operation inside a network and further improves the matching scores by leveraging a neighborhood consistency score, which is obtained by a 4D convolution layer. 
Limited by the available memory, NCNet computes the correlation scores on feature maps with 16-times downscaled resolution, which has been proven not accurate enough for camera pose estimation~\cite{Zhou2020ICRA}.
SparseNCNet~\cite{Rocco2020ECCV} uses a sparse representation of the correlation tensor by storing the top-10 similarity scores and replace dense 4D convolution with sparse convolutions. 
This allows SparseNCNet to obtain matches at 4-times downscaled resolution \wrt the original image.
DualRC-Net~\cite{Li20NIPS}, developed concurrently with our approach, outperforms SparseNCNet by combining the matching scores obtained from coarse-resolution and fine-resolution feature maps. 
Instead of refining the matching scores as in~\cite{Rocco2020ECCV, Li20NIPS}, we use regression layers to refine the match locations at image resolution.

\PAR{Full versus Weak Supervision.}
We consider methods that require information about exact correspondences to compute their loss function as fully supervised and those that do not need GT correspondences as weakly supervised.
Most local feature detectors and descriptors are trained on exact correspondences either calculated using camera poses and depth maps~\cite{Luo2020CVPR, Dusmanu2019CVPR, Germain2020ECCV} or using synthetic homography transformations~\cite{Revaud2019NIPS, Detone2018CVPRW}, except for CAPS~\cite{Wang2020ECCV} using epipolar geometry as weak supervision.
Both S2DNet~\cite{Germain2020ECCV} and SuperGlue~\cite{Sarlin2020CVPR} requires GT correspondences to learn feature description and matching.
Outlier filtering methods~\cite{Moo2018CVPR, Zhang2019ICCV, Sun2020CVPR, Brachmann2019ICCV} are normally weakly supervised by the geometry transformations between the pair. 
% \qj{
DualRC-Net~\cite{Li20NIPS} is also fully supervised on exact correspondences, while the other two correspondence networks~\cite{Rocco2018NIPS, Rocco2020ECCV} are weakly-supervised to optimize the mean matching score on the level of image pairs instead of individual matches. %}
% Both correspondence networks~\cite{Rocco2018NIPS, Rocco2020ECCV} are weakly-supervised to optimize the mean matching score on the level of image pairs instead of individual matches.
We use epipolar geometry as weak supervision to learn geometrically consistent correspondences where the coordinates of matches are directly regressed and optimize.
In contrast, CAPS ~\cite{Wang2020ECCV} uses the same level of supervision to learn feature descriptors and their loss optimizes through the matching scores whose indices give the match locations.
% To this end, w
We propose our two-stage matching network, based on the concept of learned correspondences~% as in ~
\cite{Rocco2018NIPS, Rocco2020ECCV}, which learns to predict geometrically consistent matches at image resolution.
\section{Patch2Pix: Match Refinement Network}
\begin{figure*}[t]
\centering
  \includegraphics[width=0.8\textwidth]{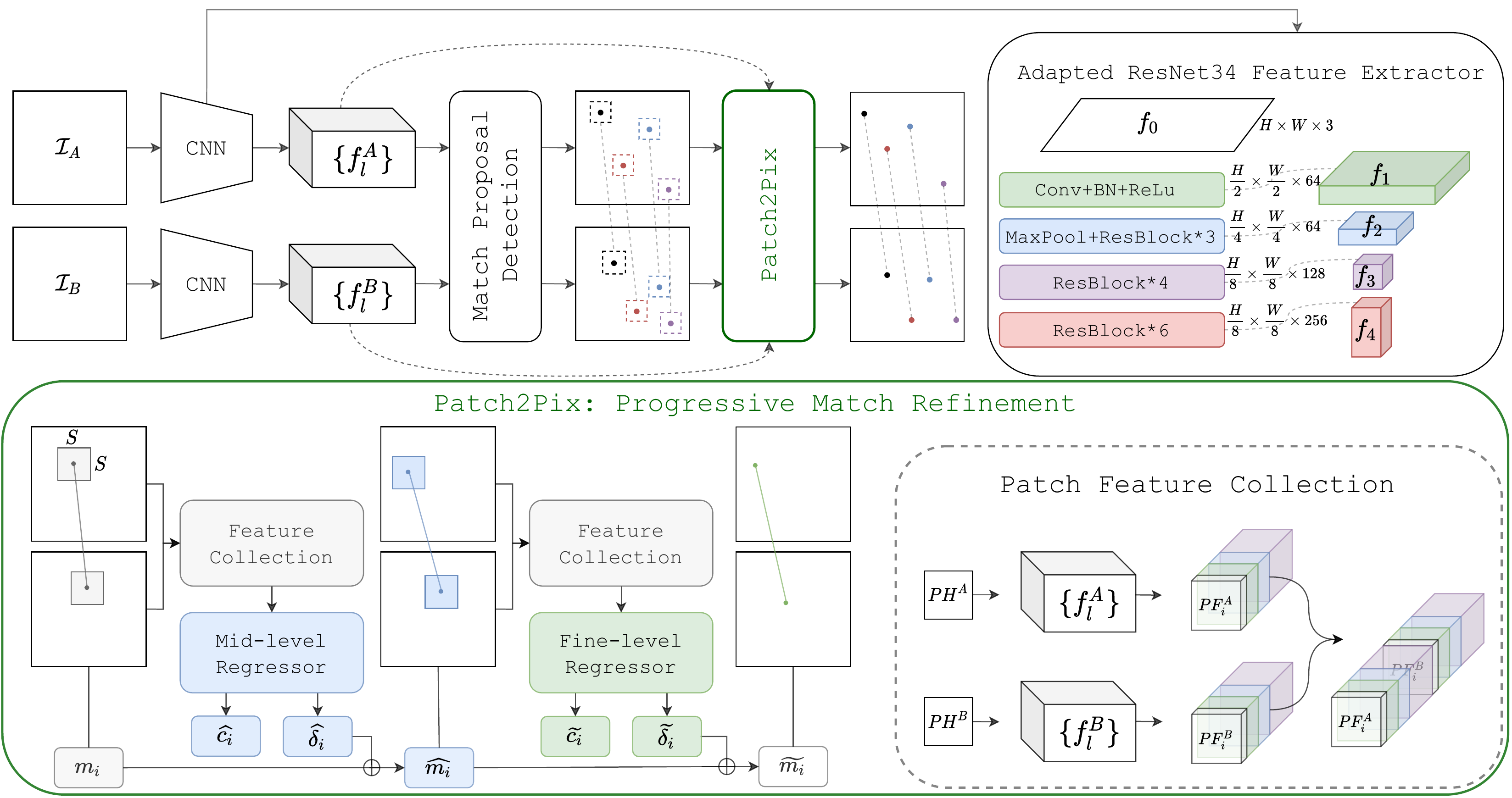}
  \caption{\textbf{Correspondence Refinement with {\it Patch2Pix}.}  {\it Top:} For a pair of images, features are first extracted using our adapted ResNet34 backbone and fed into a correspondence network, \eg, NC matching layer~\cite{Rocco2018NIPS}, to detect match proposals. Those proposals are then refined by {\it Patch2Pix}, which re-uses the extracted feature maps. {\it Bottom:} We design two levels of regressors with the same architecture to progressively refine the match proposals at image resolution. For a pair of $S\times S$ local patches centered at a match proposal $m_i$, the features of the patches are collected as the input to our mid-level regressor to output (i) a confidence score $\widehat{c_i}$ which indicates the quality of the match proposal and (ii) a pixel-level local match $\widehat{\delta_i}$ found within the local patches. The updated match proposal $\widehat{m_i}$ updates the search space accordingly through a new pair of local patches. The fine-level regressor outputs the final confidence score $\widetilde{c_i}$ and $\widetilde{\delta_i}$ to obtained the final pixel-accurate match $\widetilde{m_i}$. The whole network is trained under weak supervision without the need for explicit GT correspondences.}
  \label{fig:patch2pix}
\end{figure*}
A benefit of correspondence networks is the potential to optimize the network directly for the feature matching objective without the need for explicitly defining keypoints.
The feature detection and description are implicitly performed by the network and reflected in the found correspondences.
However, there are two main issues causing the inaccuracy of the existing correspondence networks~\cite{Rocco2020ECCV, Rocco2018NIPS}: i) the use of downscaled feature maps due to the memory bottleneck constrained by the size of the correlation map.
This leads to every match being uncertain within two local patches.
ii) Both NCNet~\cite{Rocco2018NIPS} and SparseNCNet~\cite{Rocco2020ECCV} have been trained with a weakly supervised loss which simply gives low scores for all matches of a non-matching pair and high scores for matches of a matching pair. This does not help identify good or bad matches, making the method unsuitable to locate pixel-accurate correspondences.

In order to fix those two sources of inaccuracies, we propose to perform matching in a two-stage {\it detect-to-refine} manner, which is inspired by two-step object detectors such as Faster R-CNN \cite{Ren2015NIPS}.
In the first correspondence detection stage, we adopt a correspondence network, \eg, NCNet, to predict a set of patch-level match proposals. 
As in Faster R-CNN, our second stage refines a match proposal in two ways: 
(i) using classification to identify whether a proposal is confident or not, and (ii) using regression to detect a match at pixel resolution within the local patches centered by the proposed match.
Our intuition is that the correspondence network uses the high-level features to predict semantic matches at a patch-level, while our refinement network can focus on the details of the local structure to define more accurate locations for the correspondences.
Finally, our network is trained with our weakly-supervised epipolar loss which enforces our matches to fulfill this geometric constraint defined by the relative camera pose.
We name our network \textit{Patch2Pix} since it predicts pixel-level matches from local patches, and the overview of the network architecture is depicted in Fig.~\ref{fig:patch2pix}.  
In the following, we take NCNet as our baseline to obtain match proposals,
yet we are not limited to correspondence networks %only 
to perform the match detection.
We show later in our experiments that our refinement network also generalizes to other types of matching methods (\cf Sec.~\ref{sec:exp_aachen} \& ~\ref{sec:exp_inloc}).
The following sections detail its architecture and training losses. % in the following sections.

\subsection{Refinement: Pixel-level Matching}

\PAR{Feature Extraction.}
Given a pair of images $(I_A, I_B)$, a CNN backbone with $L$ layers extracts the feature maps from each image.
We consider $\{f^A_1\}_{l=0}^{L}$ and $\{f^B_l\}_{l=0}^{L}$ to be the activation maps at layer $l$ for images $I_A$ and $I_B$, respectively. 
At the layer index $l=0$, the feature map is the input image itself, \ie, $f^A_0 = I_A$ and $f^B_0 = I_B$. 
For an image with spatial resolution $H\times W$, the spatial dimension of feature map $f_l$ is $H/2^{l} \times W/2^{l}$ for $l \in [0, L-1]$.
For the last layer, we set the convolution stride as 1 to prevent losing too much resolution.
The feature maps are extracted once and used in both the correspondence detection and refinement stages.
The detection stage uses only the last layer features which contain more high-level information, while the refinement stage uses the features before the last layer, which contain more low-level details.

\PAR{From match proposals to patches.}
Given a match proposal $m_i=(p^A_i, p^B_i)=(x^A_i, y^A_i, x^B_i, y^B_i)$, the goal of our refinement stage is to find accurate matches on the pixel level by searching for a pixel-wise match inside %defined 
local regions. 
As the proposals were matched on a downscaled feature map, an error by one pixel in the feature map leads to inaccuracy of $2^{L-1}$ pixels in the images. 
Therefore, we define the search region as the $S \times S$ local patches centered at $p^A_i$ and $p^B_i$, where we consider $S > 2^{L-1}$ to cover a larger region than the original $2^{L-1} \times 2^{L-1}$ local patches.
Once we obtain a set of local patch pairs for all match proposals, the pixel-level matches are regressed by our network from the feature maps of the local patch pairs. 
We describe each component in detail below. % in the following paragraphs.

\begin{figure}[t]
\centering
  \includegraphics[width=0.5\textwidth]{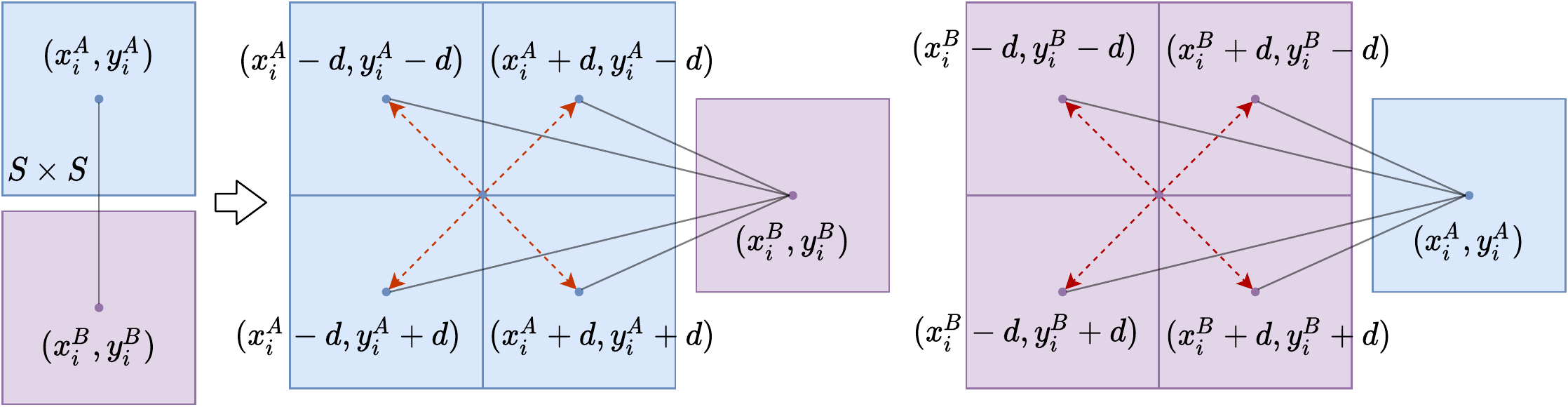}
  \caption{\textbf{Patch Expansion.} Given a match proposal $p^A_i=(x^A_i, y^A_i)$ and $p^B_i=(x^B_i, y^B_i)$, we move $p^A_i$ towards its four corners by moving along the x- and y-axes by $d$ pixels, which are matched to $p^B_i$ to compose 4 new match proposals. Repeating it also from $p^B_i$ to $p^A_i$, leads to 8 match proposals in total, which allows us to search in two $2S \times 2S $ local regions, compared to the original $S \times S$ patches.}  
  \label{fig:patch_expansion}
\end{figure}
\PAR{Local Patch Expansion.}
We further propose a patch expansion mechanism to expand the search region by including the neighboring regions, as illustrated in \figref{fig:patch_expansion}. 
We first move $p^A_i$ towards its four corners along the x- and y-axes, each by $d$ pixels. This gives us four anchor points for $p^A_i$ that we match to $p^B_i$ to compose four new match proposals. 
Similarly, we also expand $p^B_i$ to get its four corner anchors and match them to $p^A_i$, giving us another four new match proposals.
In the end, the expanded eight proposals identify eight pairs of $S \times S$ local patches.
We %specifically 
set $d=S/2$ pixels so that the expanded search region defined by the expanded patches has size $2S \times 2S$ and still covers the original $S \times S$ searching space. 
The patch expansion to the patch proposals $M_{patch}$ is especially useful during training since the network is forced to identify the correct proposal among spatially close and similar features.%, and to further get proper matches for each pair.
We show in the supp. mat (Sec.\ref{sec:supp:train_ablat}) that our expansion mechanism can speed up the learning process and also improves the model performance. 
While one can also apply it during the inference to increase the search region, it will lead to a higher computation overhead. 
We thus refrain from using it during testing.

\PAR{Progressive Match Regression.}
In order to locate pixel-level matches, we define the refinement task as finding a good match inside the pair of local patches.
We achieve this using two regressors with the same architecture, \ie, the mid-level and the fine-level regressor, to progressively identify the final match, which is shown in the lower part of ~\figref{fig:patch2pix}.
Given a pair of $S \times S$ patches, we first collect the corresponding feature information from previously extracted activation maps, \ie, $\{f^A_l\},\{f^B_l\}$.
For every point location $(x,y)$ on the patch, its corresponding location on the $l$-layer feature map is $(x/2^l, y/2^l)$. 
We select all features from the layers $\{0, \dots, L-1\}$ and concatenate them into a single feature vector.
The two gathered feature patches $PF^A_i$ and $PF^B_i$ are concatenated along the feature dimension and fed into our mid-level regressor.
The regressor first aggregates the input features with two convolutional layers into a compact feature vector, which is then processed by two fully connected (fc) layers, and finally outputs our network predictions from two  heads implemented as two fc layers.
The first head is a regression head, which outputs a set of local matches  $\widehat{M}_{\Delta}:= \{\widehat{\delta_i}\}_{i = 1}^{N} \subset R^{4}$ inside the $S \times S$ local patches \wrt their center pixels, where $\widehat{\delta_i}=(\widehat{\delta x^A_i}, \widehat{\delta y^A_i}, \widehat{\delta x^B_i}, \widehat{\delta y^B_i})$.
In the second head, \ie, the classification head, we apply a sigmoid function to the outputs of the fc layer to obtain the confidence scores $\widehat{\mathcal{C}}_{pixel}=(\widehat{c_1}, \dots, \widehat{c_N})\in R^N$, which express the validity of the detected matches. This allows us to detect and discard bad match proposals that cannot deliver a good pixel-wise match.
We obtain the mid-level matches $\widehat{M}_{pixel}:=\{\widehat{m_i}\}_{i = 1}^{N}$ by adding the local matches to patch matches, \ie, $\widehat{m_i}=m_i + \widehat{\delta_i}$. %
Features are collected again for the new set of local $S \times S$ patch pairs centered by the mid-level matches and fed into the fine-level regressor, which follows the same procedure as the mid-level regression to output the final pixel-level matches $\widetilde{M}_{pixel}:=\{\widetilde{m_i}\}_{i=1}^{N}$ and the confidence scores $\widetilde{\mathcal{C}}_{pixel}=(\widetilde{c_1}, \dots, \widetilde{c_N})\in R^N$.

\subsection{Losses}\label{sec:train_loss}
Our pixel-level matching loss $\mathcal{L}_{pixel}$ involves two terms: (i) a classification loss $\mathcal{L}_{cls}$ for the confidence scores, trained to predict whether a match proposal contains a true match or not, and (ii) a geometric loss $\mathcal{L}_{geo}$ to judge the accuracy of the regressed matches. The final loss is defined as 
$\mathcal{L}_{pixel} = \alpha\mathcal{L}_{cls} + \mathcal{L}_{geo}$, where $\alpha$ is a weighting parameter to balance the two losses. 
We empirically set $\alpha=10$ based on the magnitude of the two losses during training.

\PAR{Sampson distance.} To identify pixel-level matches, we supervise the network to find correspondences that agree with the epipolar geometry between an image pair.
It defines that the two correctly matched points should lie on their corresponding epipolar lines when being projected to the other image using the relative camera pose transformation.
How much a match prediction fulfills the epipolar geometry can be precisely measured by the Sampson distance. %\footnote{We choose the Sampson distance instead of the symmetric epipolar distance since it has been shown to work better in practice~\cite{Hartley2004}.}.
Given a match $m_i$ and the fundamental matrix $F \in R^{3\times 3}$ computed by the relative camera pose of the image pair, its Sampson distance $\phi_i$ measures the geometric error of the match \wrt the fundamental matrix~\cite{Hartley2004}, which is defined as:
\begin{equation}\label{eq:sampson_loss}
%\resizebox{7.5cm}{!}{
%$\phi_i=\Phi(m_i, F) = \frac{((P^B_i)^{T}FP^A_i)^2}{(FP^A_i)_1^2 + (FP^A_i)_2^2 + (F^TP^B_i)_1^2 + (F^TP^B_i)_2^2},$
%}
\phi_i=\frac{((P^B_i)^{T}FP^A_i)^2}{(FP^A_i)_1^2 + (FP^A_i)_2^2 + (F^TP^B_i)_1^2 + (F^TP^B_i)_2^2},
\end{equation}
where $P^A_i=(x^A_i, y^A_i, 1)^T, P^B_i=(x^B_i, y^B_i, 1)^T$ and $(FP^A_i)_k^2 , (FP^B_i)_k^2$ represent the square of the $k$-th entry of the vector $FP^A_i, FP^B_i$.

\PAR{Classification loss.} Given a pair of patches obtained from a match proposal $m_i=(x^A_i, y^A_i, x^B_i, y^B_i)$, we label the pair as positive, hence define its classification label as $c_i^*=1$, if $\phi_i <\theta_{cls}$. 
Here, $\theta_{cls}$ is our geometric distance threshold for classification. All the others pairs are labeled as negative.
Given the set of predicted confidence scores $\mathcal{C}$ and the binary labels $\mathcal{C}^*$, we use the weighted binary cross entropy to measure the classification loss as \vspace{-2pt}
\begin{equation}\label{eq:bce_loss}
\mathcal{B}(\mathcal{C},  {\mathcal{C}^*}) = - \frac{1}{N}\sum_{i=1}^{N}  w{c}^*_i \, log \, c_i + (1-{c}^*_i) \, log \, (1 - c_i)\, ,
\end{equation}
where the weight $w=|\{c^*_i|c^*_i=0\}| / |\{c^*_i|c^*_i=1\}|$ is the factor to balance the amount of positive and negative patch pairs.
We have separate thresholds $\widehat{\theta}_{cls}$ and $\widetilde{\theta}_{cls}$ used in the mid-level and the fine-level classification loss, which are summed to get the total classification loss $\mathcal{L}_{cls}$.

\PAR{Geometric loss.}
To avoid training our regressors to refine matches within match proposals which are going to be classified as non-valid, for every refined match, we optimize its geometric loss only if the Sampson distance of its parent match proposal is within a certain threshold $\theta_{geo}$.
Our geometric loss is the average Sampson distance of the set of refined matches that we want to optimize.
We use thresholds $\widehat{\theta}_{geo}$ and $\widetilde{\theta}_{geo}$ for the mid-level and the fine-level geometric loss accordingly and the sum of the two losses gives the total geometric loss $\mathcal{L}_{geo}$.

%%%%%%%%%%%%%%%%%%%%%%%%%%%%%%%%%%%
\begin{figure*}
\begin{minipage}{0.65\textwidth}
  \includegraphics[width=1\textwidth]{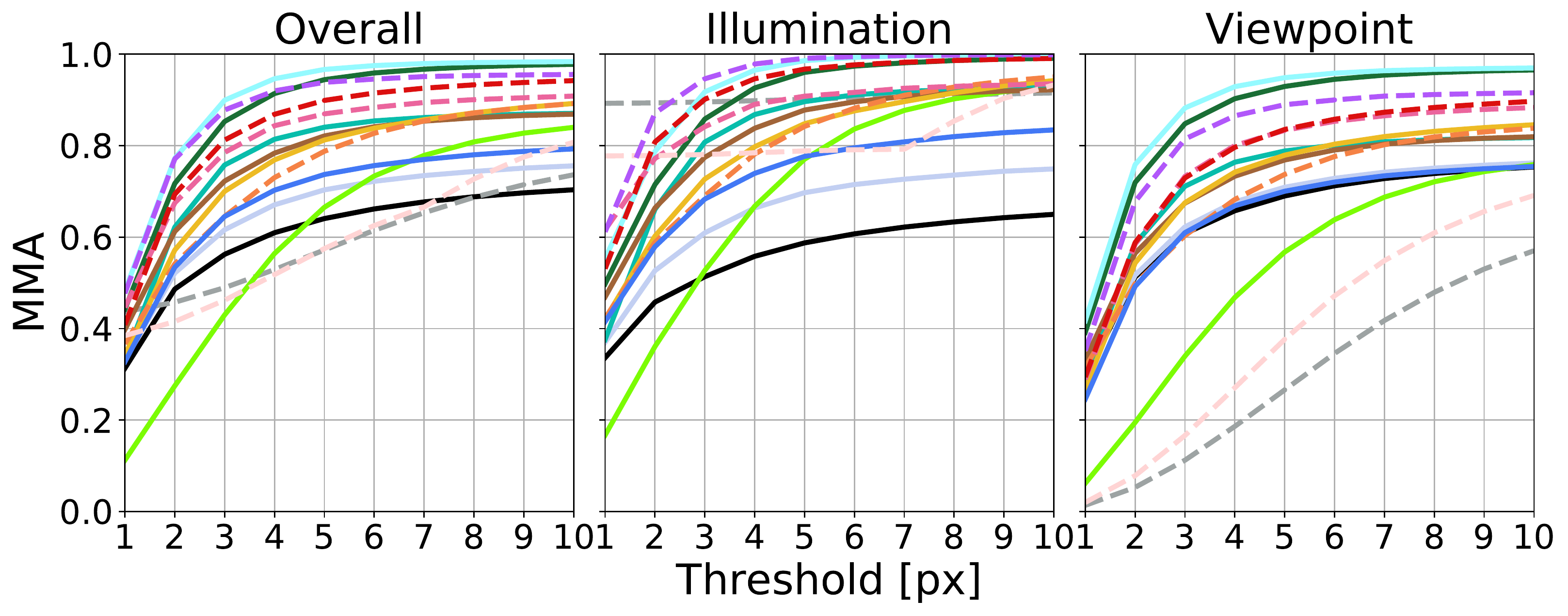}
\end{minipage}
\begin{minipage}{0.3\textwidth}
\resizebox{5cm}{!} {
\begin{tabular}{l  c }
    \toprule
	Methods  & \#Features / Matches\\
	\midrule
	\cstick{cdark} HesAff~\cite{Mikolajczyk2004IJCV} + RootSIFT + NN & 6.7K / 2.8K \\
 	\cstick{cpurblue} HAN~\cite{Mishkin2018ECCV} + HN++~\cite{Mishchuk2017NIPS} + NN &3.9K / 2.0K\\
    \cstick{cblue} SuperPoint~\cite{Detone2018CVPRW} + NN & 2.0K / 1.1K \\
    \cstick{cgrass} D2Net~\cite{Dusmanu2019CVPR} + NN & 6.0K / 2.5K  \\
    \cstick{ccyan} R2D2~\cite{Revaud2019NIPS} + NN &  5.0K / 1.6K \\	    
    \cstick{cbrown} ASLFeat~\cite{Luo2020CVPR} + NN &  4.0K / 2.0K \\
    \cstick{cgreen} SuperPoint + SuperGlue~\cite{Sarlin2020CVPR} (c=0.2) & 0.5K  \\
    \cstick{csky} SuperPoint + SuperGlue~\cite{Sarlin2020CVPR} (c=0.9) & 0.4K \\
    \cstick{cyellow} SuperPoint +  CAPS~\cite{Wang2020ECCV} + NN &  2.0K / 1.1K \\
    \midrule    
    \cdash{corange} SIFT~\cite{Lowe2004IJCV} + CAPS + NN &  4.4K / 1.5K \\
	\cdash{cgrey} DELF~\cite{Noh2017ICCV} + NN & 4.6K / 1.9K \\
    \cdash{crose} SparseNCNet~\cite{Rocco2020ECCV} (im3200, top2k) & 2.0K \\
    \cdash{cpink} NCNet(~\cite{Rocco2018NIPS} (Our Adapted) & 1.5K \\	
    \cdash{cred} Patch2Pix(c=0.5) & 1.1K  \\ 	
	\cdash{cpurple} Patch2Pix(c=0.9) & 0.7K \\
\bottomrule	
\end{tabular}
}
\end{minipage}
\caption{\textbf{Image Matching on HPatches~\cite{Balntas2017CVPR}.} We denote weakly-supervised methods with dashed lines and methods based on full supervision with solid lines.}
\label{fig:exp_matching}
\end{figure*}

\section{Implementation Details}\label{sec:implementation}
% \PAR{Training details.} 
We train \textit{Patch2Pix} with match proposals detected by our adapted NCNet, \ie, the pre-trained NC matching layer from~\cite{Rocco2018NIPS}, to match features extracted from our backbone.
Our refinement network is trained on the large-scale outdoor dataset MegaDepth~\cite{Li18CVPR}, where we construct 60661 matching pairs.
We set the distance thresholds to compute the training losses  (\cf Sec.~\ref{sec:train_loss}) as $\widehat\theta_{cls}=\widehat\theta_{geo}=50$ for the mid-level regression and $\widetilde{\theta}_{cls}=\widetilde{\theta}_{geo}=5$ for the fine-level regression. 
We constantly set the local patch size to $S=16$ pixels at image resolution. 
%to be 2-times larger than the downscale size of the last feature map \wrt image resolution.
The pixel-level matching is optimized using Adam~\cite{Diederik2015ICLR} with an initial learning rate of $5e^{-4}$ for 5 epochs and then $1e^{-4}$ until it converges.
A mini-batch input contains 4 pairs of images with resolution $480 \times 320$.
We present architecture details about our regressor and our adapted NCNet~\cite{Rocco2018NIPS}, training data processing, hyper-parameter ablation, and qualitatively results of our matches in the supp. mat. (\cf Sec.\ref{sec:supp:implement_details} \& Sec.\ref{sec:supp:train_ablat}).

\section{Evaluation on Geometrical Tasks}
\subsection{Image Matching}\label{sec:exp_matching}
As our first experiment, we evaluate {\it Patch2Pix} on the HPatches~\cite{Balntas2017CVPR} sequences under the image matching task, where a method is supposed to detect correspondences between an input image pair.
We follow the setup proposed in D2Net~\cite{Dusmanu2019CVPR} and report the mean matching accuracy (MMA)~\cite{Mikolajczyk2005TPAMI} under thresholds varying from 1 to 10 pixels, together with the numbers of matches and features.

\PAR{Experimental setup.} 
We use the confidence scores produced by the fine-level regressor to filter out outliers and study its performance under two settings, \ie, $c=0.5/0.9$, which present a trade-off between quantity and quality of the matches. 
To show the effectiveness of our refinement concept, we compare to our NCNet baseline, which provides our match proposals.
For NCNet and {\it Patch2Pix}, we resize images to have a larger side of 1024 to reduce runtime.
We also compare to SparseNCNet~\cite{Rocco2020ECCV}, which is the most similar one to ours among related works, since it also builds upon NCNet and aims to improve the accuracy of its matches through a re-localization mechanism.
Besides comparing to several local feature methods that use NN Search for matching, we further consider SuperPoint~\cite{Detone2018CVPRW} features matched with SuperGlue~\cite{Sarlin2020CVPR} and study its performance under their default threshold $c=0.2$ and a higher threshold  $c=0.9$ for outlier rejection.

\PAR{Results.} 
As shown in Fig.~\ref{fig:exp_matching}, NCNet performs competitively for illumination sequences with constant viewpoints, which is a special case for NCNet since it uses fixed upsampling to bring patch matches to pixel correspondences. 
While its performance under illumination changes reveals its efficiency in patch-level matching, its accuracy under viewpoint changes reveals its insufficient pixel-level matching performance.
% We show that o
Our refinement network brings patch-level matches predicted by NCNet to pixel-level correspondences, which drastically improves the matching accuracy under viewpoint changes and further improves under illumination changes.
When comparing {\it Patch2Pix} to all weakly supervised methods, our model is the best at both thresholds under illumination changes.
For viewpoint changes, our model with threshold $c=0.9$ is the best and SparseNCNet performs similar to our model under threshold $c=0.5$.
Compared to the methods trained with full supervision, our model with threshold $c=0.9$ outperforms all of them under illumination variations.
For viewpoint changes, we are less accurate than SuperPoint + SuperGlue but still, we outperform all the other fully-supervised methods.
Looking at the curves and the table in ~\figref{fig:exp_matching} together, %we see 
both SuperPoint + SuperGlue and our method improve performance when using a higher threshold to remove less confident predictions.

%%%%%%%%%
\subsection{Homography Estimation} \label{sec:exp_homography}
\begin{table*}[t]
\centering
\setlength{\tabcolsep}{8pt}
\resizebox{16cm}{!} {
\begin{tabular}{l | c c c | c  c  c  c }
    \toprule
	\multirow{2}{*}{Method} &  Overall & Illumination & Viewpoint & \multirow{2}{*}{Supervision} & \multirow{2}{*}{\#Matches} & \multirow{2}{*}{Time (s)} \\
	  & \multicolumn{3}{c|}{Accuracy ($\%,\epsilon< 1/3/5$ px)} & & & &\\
	\midrule
    SuperPoint~\cite{Detone2018CVPRW} + NN  & 0.46 / 0.78 / 0.85 & 0.57 / 0.92 / 0.97 & 0.35 / 0.65 / 0.74  & Full & 1.1K &  0.12 \\
	D2Net~\cite{Dusmanu2019CVPR} + NN       & 0.38 / 0.72 / 0.81 & 0.65 / 0.95 / \textbf{0.98} & 0.13 / 0.51 / 0.65  & Full & 2.5K & 1.61 \\
	R2D2~\cite{Revaud2019NIPS} + NN         & 0.47 / 0.78 / 0.83 & 0.63 / 0.93 / \textbf{0.98} & 0.33 / 0.64 / 0.70  & Full & 1.6K & 2.34 \\
	ASLFeat~\cite{Luo2020CVPR} + NN & 0.48 / 0.81 / 0.88 & 0.63 / 0.94 / \textbf{0.98} & 0.34 / 0.69 / 0.78   & Full & 2.0K & 0.66 \\
	SuperPoint + SuperGlue~\cite{Sarlin2020CVPR}
	                                        &  \textbf{0.51 / 0.83 / 0.89} & 0.62 / 0.93 / \textbf{0.98} & \textbf{0.41/ 0.73/ 0.81} & Full & 0.5K & 0.14 \\
    SuperPoint + CAPS~\cite{Wang2020ECCV} + NN& 0.49 / 0.79 / 0.86 & 0.62 / 0.93 / 0.98  & 0.36 / 0.65 / 0.75  & Mix & 1.1K & 0.36 \\
    \midrule	                                        
    SIFT + CAPS~\cite{Wang2020ECCV} + NN    & 0.36 / 0.76 / 0.85 & 0.48 / 0.89 / 0.95 & 0.26 / 0.65 / 0.76 & Weak & 1.5K & 0.73 \\
    SparseNCNet~\cite{Rocco2020ECCV} (im3200, top2k) & 0.36 / 0.66 / 0.76 & 0.62 / 0.92 / 0.97 & 0.13 / 0.41 / 0.57 & Weak & 2.0K & 5.83\\
    NCNet~\cite{Rocco2018NIPS} (Our Adapted)    & 0.48 / 0.61 / 0.71 & \textbf{0.98 / 0.98 / 0.98} & 0.02 / 0.28 / 0.46 & Weak & 1.5K & 0.83 \\
    Patch2Pix                                & \textbf{0.51} / 0.79 / 0.86  & 0.72 / 0.95 / \textbf{0.98} & 0.32 / 0.64 / 0.75 & Weak & 1.3K & 1.24 \\
    \midrule \midrule
    Oracle               & 0.00 / 0.15  / 0.54  & 0.00 / 0.23 / 0.7& 0.00 / 0.07 / 0.39 & - & 2.5K & 0.04 \\
    Patch2Pix (w.Oracle)        & 0.55 / 0.85 / 0.92 & 0.68 / 0.95 / 0.99 & 0.43 / 0.76 / 0.82 & Weak & 2.5K & 0.76 \\  
	\bottomrule	
\end{tabular}
}
 \vspace{3pt}
\caption{\textbf{Homography Estimation on Hpatches~\cite{Balntas2017CVPR}.} We report the percentage of correctly estimated homographies whose average corner error distance is below 1/3/5 pixels. We denote the supervision type with 'Full' for fully-supervised methods, 'Weak' for weakly-supervised ones, and 'Mix' for those used both types. We mark the best accuracy in \textbf{bold}.}
\label{tab:exp_homography}

\end{table*}

Having accurate matches does not necessarily mean accurate geometry relations can be estimated from them since the distribution and number of matches are also important when estimating geometric relations. % the existing geometry solvers. 
Therefore, we next evaluate {\it Patch2Pix} on the same HPatches~\cite{Balntas2017CVPR} sequences for homography estimation.

\PAR{Experimental setup.} 
We follow the corner correctness metric used in ~\cite{Detone2018CVPRW, Sarlin2020CVPR, Wang2020ECCV} and report the percentage of correctly estimated homographies whose average corner error distance is below 1/3/5 pixels.
In the following experiments, where geometrics relations are estimated using RANSAC-based solvers, we use $c=0.25$ as our default confidence threshold, which overall gives us good performance across tasks.
The intuition of setting a lower threshold is to filter out some very bad matches but leave as much information as possible for RANSAC to do its own outlier rejection.
We compare to methods that are more competitive in the matching task which are categorized based on their supervision types: fully supervised (Full), weakly supervised (Weak), and mixed (Mix) if both types are used.
We run all methods under our environment and measure the matching time from the input images to the output matches.
We provide more experimental setup details in our supp. mat (\cf Sec.\ref{sec:supp:exp_details}).

\PAR{Results.}
From the results shown in Tab.~\ref{tab:exp_homography}, we observe again that NCNet performs extremely well under illumination changes due to their fixed upsampling (\cf Sec.~\ref{sec:exp_homography}).
Here, we verify that the improvement of matches by {\it Patch2Pix} under viewpoint changes is also reflected in the quality of the estimated homographies. % estimated by us.
Both SparseNCNet and our method are based on the concept of improving match accuracy by searching inside the matched local patches to progressively re-locate a more accurate match in higher resolution feature maps. While our method predicts matches at the original resolution and is fully learnable, their non-learning approach produces matches at a 4-times downscaled resolution.
As we show in Tab.~\ref{tab:exp_homography}, our refinement network is more powerful than their re-localization mechanism, improving the overall accuracy within 1 pixel by 15 percent.
For illumination changes, we are the second-best after NCNet, but we are better than all fully supervised methods.
Under viewpoint variations, we are the best at 1-pixel error among weakly-supervised methods and we achieve very close overall accuracy to the best fully supervised method SuperPoint + SuperGlue.

\PAR{Oracle Investigation.}
Since our method can filter out bad proposals but not generate new ones, our performance will suffer if NCNet fails to produce enough valid proposals, which might be the reason for our relatively lower performance on viewpoint changes.
In order to test our hypothesis, we replace NCNet with an Oracle matcher to predict match proposals.
Given a pair of images, our Oracle first random selects 2.5K matches from the GT correspondences computed using the GT homography and then randomly moves each point involved in a match within the $12 \times 12$ local patch centered at the GT location.
In this way, we obtain our synthetic match proposals where we know there exists at least one GT correspondence inside the $16 \times 16$ local patches centered by those match proposals, which allows us to measure the performance of our true contribution, the refinement network.
As shown in Tab.~\ref{tab:exp_homography}, the low accuracy of matches produced by our Oracle evidently verifies that the matching task left for our refinement network is still challenging. 
Our results are largely improved by using the Oracle proposals, which means our current refinement network is heavily limited by the performance of NCNet.
Therefore, in the following localization experiments, to see the potential of our refinement network, we will also investigate the performance when using SuperPoint + SuperGlue to generate match proposals.

%%%%%%%%%%%%%%%Localization%%%%%%%%%%%%

\subsection{Outdoor Localization on Aachen Day-Night}\label{sec:exp_aachen}
We further show the potential of our approach by evaluating {\it Patch2Pix} on the Aachen Day-Night benchmark (v1.0)~\cite{Sattler2018CVPR,Sattler2012BMVC} for outdoor localization under day-night illumination changes.
\begin{table}[t]
\centering
\setlength{\tabcolsep}{4pt}
\resizebox{8.5cm}{!} {
\begin{tabular}{l c c c}
    \toprule
     \multirow{2}{*}{Method}& \multirow{2}{*}{Supervision}& \multicolumn{2}{c}{Localized Queries (\%, 0.25$m$,2$^\circ$/0.5$m$,5$^\circ$/1.0$m$, 10$^\circ$)}\\
    & &  Day & Night\\
	\midrule
    \multicolumn{3}{l}{\textbf{Local Feature Evaluation on Night-time Queries}} & \\
    \midrule
    SuperPoint~\cite{Detone2018CVPRW} + NN      & Full & - & 73.5 / 79.6 / 88.8 \\    
    D2Net~\cite{Dusmanu2019CVPR} + NN           & Full & - & 74.5 / 86.7 / \textbf{100.0} \\
    R2D2~\cite{Revaud2019NIPS} + NN             & Full & - & 76.5 /\textbf{ 90.8} / \textbf{100.0} \\
    SuperPoint + S2DNet~\cite{Germain2020ECCV}  & Full & - & 74.5 / 84.7 / \textbf{100.0} \\
    % ASLFeat(v2)~\cite{Luo2020CVPR} + NN         & Full & - & 81.6 / 87.8 / 100.0 \\
    ASLFeat~\cite{Luo2020CVPR} + NN         	& Full & - & 77.6 / 89.8 / \textbf{100.0} \\
    SuperPoint + CAPS~\cite{Wang2020ECCV} + NN  & Mix & - & \textbf{82.7} / 87.8 / \textbf{100.0} \\ 
    DualRC-Net~\cite{Li20NIPS}  & Full & - & 79.6 / 88.8 / 100.0 \\
    \midrule        
    SIFT + CAPS~\cite{Wang2020ECCV} + NN        & Weak & - & 77.6 / 86.7 / 99.0 \\
    SparseNCNet~\cite{Rocco2020ECCV}  			& Weak & - & 76.5 / 84.7 / 98.0 \\  % (im3200, top8k)
    Patch2Pix 						   			& Weak & - & 79.6 / 87.8 / \textbf{100.0} \\     
    \midrule    
    \midrule    
	\multicolumn{2}{l}{\textbf{Full Localization with HLOC~\cite{Sarlin2019CVPR}}} & & \\
	\midrule
    SuperPoint~\cite{Detone2018CVPRW} + NN       &  Full & 85.4 / 93.3 / 97.2 & 75.5 / 86.7 / 92.9 \\
    SuperPoint + CAPS~\cite{Wang2020ECCV} + NN   &  Mix  & 86.3 / 93.0 / 95.9 &	83.7 / 90.8 / 96.9 \\
    SuperPoint + SuperGlue~\cite{Sarlin2020CVPR} &  Full & \textbf{89.6} / 95.4 / \textbf{98.8} & 86.7 / 93.9 / \textbf{100.0} \\    
    Patch2Pix                    	 &  Weak &  84.6 / 92.1 / 96.5 & 82.7 / 92.9 / 99.0 \\
    \midrule 
    Patch2Pix (w.CAPS) 				 & Mix & 86.7 / 93.7 / 96.7 & 85.7 / 92.9 / 99.0\\    
    Patch2Pix (w.SuperGlue)       	 & Mix & 89.2 / \textbf{95.5} / 98.5  & \textbf{87.8 / 94.9 / 100.0} \\ % c0.2
	\bottomrule	

\end{tabular}
}
 \vspace{3pt}
\caption{\textbf{Evaluation on Aachen Day-Night Benchmark (v1.0)~\cite{Sattler2018CVPR,Sattler2012BMVC}.} We report the percentage of correctly localized queries under specific error thresholds. We follow the supervision notations described in Tab.~\ref{tab:exp_homography} and mark the best results in \textbf{bold}.}
\label{tab:exp_aachen}
\end{table}

\PAR{Experimental Setup.} 
To localize Aachen night-time queries, we follow the evaluation setup from the website\footnote{https://github.com/tsattler/visuallocalizationbenchmark}.
For evaluation on day-time and night-time images together, we adopt the hierarchical localization pipeline  (HLOC\footnote{https://github.com/cvg/Hierarchical-Localization}) proposed in ~\cite{Sarlin2019CVPR}.
Matching methods are then plugged into the pipeline to estimate 2D correspondences. 
We report the percentage of correctly localized queries under specific error thresholds.
We test our {\it Patch2Pix} model with NCNet proposals and SuperPoint~\cite{Detone2018CVPRW} + SuperGlue~\cite{Sarlin2020CVPR} proposals. Note, the model has been only trained on NCNet proposals.
Due to the triangulation stage inside the localization pipeline, we quantize our matches by representing keypoints that are closer than 4 pixels to each other with their mean location. We provide a more detailed discussion of the quantization inside our supp. mat (\cf Sec.\ref{sec:supp:exp_details}).

\PAR{Results.} 
As shown in Tab.~\ref{tab:exp_aachen}, for local feature evaluation on night-time queries, we outperform the other two weakly-supervised methods.
While being worse than SuperPoint~\cite{Detone2018CVPRW} + CAPS~\cite{Wang2020ECCV}, which involves both full and weak supervision, we are on-par or better than all the other fully-supervised methods.
For full localization on all queries using HLOC, we show we are better than SuperPoint + NN on night queries and competitively on day-time images.
By further substituting NCNet match proposals with SuperGlue proposals, we are competitive to SuperGlue on day-time images and outperform them slightly on night queries.
Our intuition is that we benefit from our epipolar geometry supervision which learns potentially more general features without having any bias from the training data, which is further supported by our next experiment.

%%%%%%%%%%%%%%%%%%%%%%
\begin{table}[t]
\centering
\setlength{\tabcolsep}{4pt}
%\scriptsize{
\resizebox{8.5cm}{!} {
\begin{tabular}{l c c c}
    \toprule
    \multirow{2}{*}{Method}& \multirow{2}{*}{Supervision}& \multicolumn{2}{c}{Localized Queries (\%, 0.25$m$/0.5$m$/1.0$m$, 10$^\circ$)} \\
    &  & DUC1 & DUC2 \\
    \midrule
    SuperPoint~\cite{Detone2018CVPRW} + NN   & Full &  40.4 / 58.1 / 69.7  & 42.0 / 58.8 / 69.5   \\
    D2Net~\cite{Dusmanu2019CVPR} + NN  & Full & 38.4 / 56.1 / 71.2  &  37.4 / 55.0 / 64.9\\ 
    R2D2~\cite{Revaud2019NIPS} + NN  & Full & 36.4 / 57.6 / 74.2 & 45.0 / 60.3 / 67.9 \\    
    SuperPoint + SuperGlue~\cite{Sarlin2020CVPR}& Full & 49.0 / \textbf{68.7} / 80.8 & 53.4 / 77.1 / \textbf{82.4} \\ 
    SuperPoint + CAPS~\cite{Wang2020ECCV} + NN  & Mix & 40.9 / 60.6 / 72.7 &  43.5 / 58.8 / 68.7 \\ 
    \midrule
    SIFT + CAPS~\cite{Wang2020ECCV} + NN 	& Weak & 38.4 / 56.6 / 70.7 & 35.1 / 48.9 / 58.8 \\
    SparseNCNet~\cite{Rocco2020ECCV} 					& Weak & 41.9 / 62.1 / 72.7 & 35.1 / 48.1 / 55.0 \\
    Patch2Pix     			         & Weak & 44.4 / 66.7 / 78.3  &  49.6 / 64.9 / 72.5     \\    
    \midrule 
    % % \midrule 
	Patch2Pix (w.SuperPoint+CAPS)   & Mix & 42.4 / 62.6 / 76.3	& 43.5 / 61.1 / 71.0 \\    
    Patch2Pix (w.SuperGlue)      & Mix & \textbf{50.0} / 68.2 / \textbf{81.8} &  \textbf{57.3 / 77.9 }/ 80.2 \\  % c0.1
    \bottomrule	
\end{tabular}
}
 \vspace{3pt}
\caption{\textbf{InLoc~\cite{Taira2018CVPR} Benchmark Results.} We report the percentage of correctly localized queries under specific error thresholds. Methods are evaluated inside the HLOC~\cite{Sarlin2019CVPR} pipeline to share the same retrieval pairs, RANSAC threshold, \etc We use the supervision notation from  Tab.~\ref{tab:exp_homography} and mark the best results in \textbf{bold}.}
\label{tab:exp_inloc}
\end{table}

\subsection{Indoor Localization on InLoc } \label{sec:exp_inloc}
Finally, we evaluate {\it Patch2Pix} on the InLoc benchmark ~\cite{Taira2018CVPR} for large-scale indoor localization.
The large texture-less areas and repetitive structures present in its scenes makes this dataset very challenging.
\PAR{Experimental Setup.} 
Following SuperGlue~\cite{Sarlin2020CVPR}, we evaluate a matching method by using their predicted correspondences inside HLOC for localization.
We report the percentage of correctly localized queries under specific error thresholds.
It is worth noting that compared to the evaluation on Aachen Day-Night, where our method looses accuracy up to 4 pixels due to the quantization, we have a fairer comparison on InLoc (where no triangulation is needed) to other methods. 
The results directly reflect the effect of our refinement  when combined with other methods.
Except for SuperPoint+SuperGlue, we evaluate several configurations of the other methods and compare to their best results.
Please see the supp. mat. for more details (\cf Sec.\ref{sec:supp:exp_details}).

\PAR{Results.}
As shown in Tab.~\ref{tab:exp_inloc},  {\it Patch2Pix} is the best among weakly supervised methods and  outperforms all other methods except for SuperPoint + SuperGlue.
Notice, we are 14.5 \% better than SparseNCNet on DUC2 at the finest error, which further highlights that our learned refinement network is more effective than their hand-crafted relocalization mechanism. 
Further looking at the last rows of Tab.~\ref{tab:exp_inloc}, our refinement network achieves the overall best performance among all methods when we replace NCNet proposals with more accurate proposals predicted by SuperPoint + SuperGlue. 
By searching inside the local regions of SuperPoint keypoints that are matched by SuperGlue, our network is able to detect more accurate and robust matches to outperform SuperPoint + SuperGlue. 
This implies that epipolar geometry is a promising type of supervision for the matching task.
While CAPS is also trained with epipolar loss, its performance still largely relies on the keypoint detection stage. 
In contrast, we bypass the keypoint detection errors by working directly on the potential matches.

\PAR{Generalization}
By evaluating {\it Patch2Pix} on image matching (\cf Sec.~\ref{sec:exp_matching}) and homography estimation  (\cf Sec.~\ref{sec:exp_homography}), we validate our refinement concept by showing dramatic improvements over NCNet matches.
While our network has been trained only on NCNet-type of proposals, we show that our refinement network provides distinct improvements, on both indoor and outdoor localization, by switching from the match proposals produced by NCNet to SuperPoint + SuperGlue proposals without the need for retraining. 
This highlights that our refinement network learns the general task of predicting matches from a pair of local patches, which works across different scene types and is independent of how the local patch pair has been obtained.
Such general matching capability can be used to further improve the existing methods.
As shown in Tab.~\ref{tab:exp_aachen} and Tab.~\ref{tab:exp_inloc}, both SuperPoint + SuperGlue and SuperPoint + CAPS get improved by using our refinement network.  

\section{Conclusion}
% Old 
% In this paper, we proposed a two-stage end-to-end matching network that takes a pair of images as input and directly predicts matches with pixel accuracy. 
% %
% By dividing the matching process in two stages, we allowed the first stage to focus on capturing the semantic high-level information while the second stage focused on the detailed structures inside the patches.
% %
% Our network was trained in a weakly supervised manner for finding geometrically consistent correspondences without the need for ground truth correspondences.
% %
% Our method achieves the best results among weakly supervised methods and performing competitively to the state-of-the-art full supervised methods on a variety of geometry tasks.
% We show our novel idea of regressing directly the location of correspondences from CNN features working surprisingly well, which provides inspiration to the future research work on the matching task.

In this paper, we proposed a new paradigm to predict correspondences in a two-stage \textit{detect-to-refine} manner, where the first stage focuses on capturing the semantic high-level information and the second stage focuses on the detailed structures inside local patches.
To investigate the potential of this concept, we developed a novel refinement network, which leverages regression to directly output the locations of matches from CNN features and jointly predict confidence scores for outlier rejection.
Our network was weakly supervised by epipolar geometry to detect geometrically consistent correspondences.% without the need for GT correspondences.
We showed that our refinement network consistently improved our correspondence network baseline on a variety of geometry tasks.
We further showed that our model trained with proposals predicted by a correspondence network generalizes well to other types of proposals during testing.
By applying our refinement to the best fully-supervised method without retraining, we achieved state-of-the-art results on challenging long-term localization tasks.
\clearpage

%%%%%%%%%%%%%%%Supplementary as Appendix%%%%%%%%%%%%%%%%%%%%%%%%
\appendix
\section*{Supplementray Material}
\noindent In this supplementary material, we provide additional information to further understand our proposed refinement network \textit{Patch2Pix}. 
In Sec.~\ref{sec:supp:implement_details}, we provide the architecture details of our backbone, regressors and our baseline, \ie, the adapted NCNet, followed by the details of our training data and other implementation details. 
We presents ablation studies on our architecture and training hyper-parameters in Sec.~\ref{sec:supp:train_ablat}. % and the learning rate. 
We further detail the experimental setups for the homography estimation and outdoor/indoor localization in Sec.~\ref{sec:supp:exp_details}.
Finally, Sec.~\ref{sec:supp:qualitative} shows qualitative results of matches estimated by \textit{Patch2Pix} on various benchmarks.
We will release our code upon the paper's acceptance.

\begin{figure}[b]
\centering
  \includegraphics[width=0.45\textwidth]{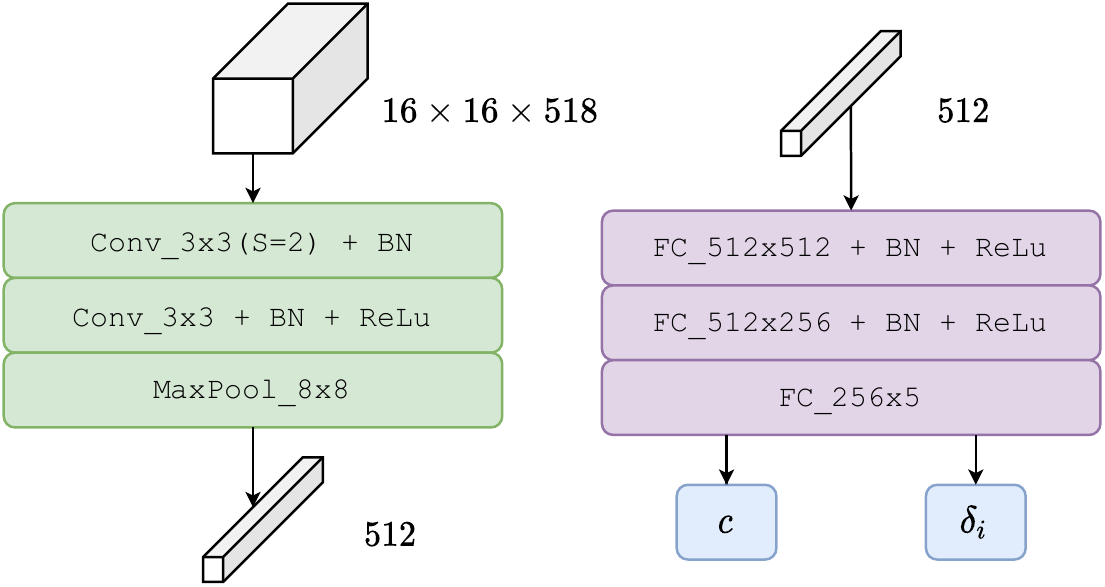}
  \caption{\textbf{Regressor Architecture.}}  
  \label{fig:regressor}
%   \vspace{-9pt}
\end{figure}

\section{Implementation Details.} \label{sec:supp:implement_details}
\PAR{Backbone.} We use a truncated ResNet34 as our backbone to extract features from the input images. It predicts 5 feature maps, from input feature $f_0$ to the last feature map $f_4$. The corresponding feature channel dimensions are [3, 64, 64, 128, 256]. 
To have enough resolution in the last feature map $f_4$, we change the stride of the convolutional layer to prevent further downscaling, which means the spatial resolution of $f_4$ is the same as $f_3$, \ie, $1/8$ of the original image resolution.
The ResNet34 backbone is pretrained on ImageNet~\cite{Deng2009CVPR} and frozen during training. 
\PAR{Regressor.} Our mid-level and fine-level regressors have the same architecture, as shown in \figref{fig:regressor}. 
On the left side of the figure, the collected features from the backbone of a patch pair are fed into a set of layers to aggregate the feature tensors into a single vector. 
On the right side, the aggregated feature vector is fed into a set of fully connected layers (FC) to output the confidence score $c$ and the coordinates of the detected local match $\delta_i$.
\PAR{Our Adapted NCNet~\cite{Rocco2018NIPS}.}
To detect match proposals, we use the pretrained NC matching layer from NCNet ~\cite{Rocco2018NIPS} to match our extracted features.
Given a pair of images, features are first extracted from the image and the two last feature maps, \ie $f^A_4, f^B_4$, which are 8-times downscaled \wrt image resolution, are exhaustively matched to produce a correlation map.
The size of the correlation map is further reduced using a MaxPool4D operation with window size $k=2$ for computational efficiency following the original NCNet~\cite{Rocco2018NIPS}.
% Notice, this step does not change the feature matching resolution but only keeps the most confident correlation score within a $k\times k \times k \times k$ local region for computational efficiency.
The final matching score map is obtained by applying a 4D convolution over the reduced correlation map to enforce neighbourhood consensus.
The raw matches are the indices of row-wise and column-wise maximum values of the matching score map.
To go back to the matching resolution, the raw matches are shifted to the corresponding pooling location using the index information from the MaxPool4D operation.
This results in downscaled matches, with each match corresponding to a pair of local $8 \times 8$ patches in the original image.
Multiplying a match by 8 gives the two upper-left corners of the two local patches.

We keep only the mutually matched patches and use all of them during training. 
During inference, we further filter the mutually matched ones with a match score threshold $c=0.9$ for outlier rejection.
We found it produces the best performance across tasks in our experiments.
% We found a large portion of their learned matching scores are above 0.9 which we think is caused by their mean matching loss.
% We pick threshold 0.9 for NCNet, which produces the best performance in our experiments.

\begin{table*}[t]
\centering
\setlength{\tabcolsep}{4pt}
\resizebox{12cm}{!} {
\begin{tabular}{c c c c c c c c }
    \toprule
	\multirow{2}{*}{ID} & \multirow{2}{*}{Exp.} & \multirow{2}{*}{Feat.} & \multirow{2}{*}{$\widehat{\theta}_{cls}$ / $\widehat{\theta}_{geo}$} &\multirow{2}{*}{$\widetilde{\theta}_{cls}$/$\widetilde{\theta}_{geo}$} &  Overall & Illumination & Viewpoint \\
	&&&&&\multicolumn{3}{c}{Accuracy ($\%,\epsilon< 1/3/5$ px)} \\
	\midrule
	\multirow{5}{*}{I} & \multirow{5}{*}{No} 
	 & 01   &\multirow{5}{*}{50/50} & \multirow{5}{*}{5/5} & 0.37 / 0.75 / 0.82 & 0.54 / 0.91 / 0.95 & 0.20 / 0.60 / 0.70 \\
    && 012  &&& 0.34 / 0.69 / 0.8 & 0.62 / 0.92 / 0.97 & 0.08 / 0.47 / 0.64 \\	
    && 123  &&& 0.37 / 0.69 / 0.81 & \hblue{0.68 / 0.93 / 0.98} & 0.09 / 0.48 / 0.66 \\	
	&& 01234&&& 0.42 / 0.76 / 0.83 & 0.62 / 0.92 / 0.97 & 0.24 / 0.60 / 0.70 \\	
    && 0123 &&& \hblue{0.45 / 0.77 / 0.85} & 0.65 /\hblue{0.93} / \hblue{0.98} & \hblue{0.27 / 0.62 / 0.73} \\	
    \midrule
    \multirow{6}{*}{II}  & 	\multirow{6}{*}{No} & \multirow{6}{*}{0123} 
      & 400/400 & 5/5  & 0.43 / 0.76 / 0.84 & 0.56 / 0.93 / 0.97 & \hcyan{0.27 / 0.62} / 0.73 \\
	&&& 100/100 & 5/5  & 0.45 / 0.76 / 0.84 & 0.67 / 0.93 / \hcyan{0.98} & 0.25 / 0.60 / 0.72 \\	
    &&& 50/50 & 15/15  & 0.44 / \hcyan{0.77 / 0.85} & \hcyan{0.68} / 0.93 / 0.97 & 0.21 / \hcyan{0.62} / \hcyan{0.74} \\ 		
    &&& 50/50 & 5/5  &  \hcyan{0.45 / 0.77 / 0.85} & 0.65 / 0.93 /  \hcyan{0.98} & \hcyan{0.27 / 0.62 }/ 0.73 \\
    &&& 50/50 & 1/1  & 0.43 / 0.76 / 0.84 & 0.62 / 0.91 / 0.97 & 0.26 / 0.61 / 0.71 \\ 	
    &&& 25/25 & 5/5  & 0.41 / \hcyan{0.77 / 0.85} & 0.61 / \hcyan{0.94 / 0.98} & 0.23 /\hcyan{ 0.62} / 0.73 \\
    \midrule    	
	\multirow{4}{*}{III}  & 	\multirow{4}{*}{Yes} & \multirow{4}{*}{0123} 
	  & 50/50 & 15/1 & 0.42 / 0.78 / 0.85  & 0.59 / \hgreen{0.95 / 0.98} & 0.27 / 0.61 / 0.73 \\
	&&& 50/50 & 5/5  & \hgreen{0.47 / 0.78 / 0.85} & \hgreen{0.65} / 0.93 / 0.97 & \hgreen{0.30 / 0.64 / 0.73} \\
    &&& 50/50 & 5/1  & 0.45 / 0.76 / 0.85 & 0.64 / 0.93 / \hgreen{0.98} & 0.28 / 0.59 / 0.72 \\  
    &&& 50/50 & 1/1  & 0.44 / 0.76 / 0.84 & 0.63 / 0.94 / \hgreen{0.98} & 0.26 / 0.60 / 0.71 \\ 
    % Converged & Yes & 0123 & 50/50 & 5/5 & 0.51 / 0.79 / 0.86 & 0.72 / 0.95 / 0.98 & 0.32 / 0.64 / 0.75 \\
	\bottomrule	
\end{tabular}
}
 \vspace{3pt}
\caption{\textbf{{\it Patch2Pix} Training Ablation.} All trained variants are evaluated on HPatches~\cite{Balntas2017CVPR} for homography estimation. We compare the models within each group and mark the best results in different colors.}
% In the final row, we present the accuracy of the best model that is trained until it converged.}
\label{tab:supp:exp_train_ablat}
\end{table*}
\PAR{Training Data Processing.}
Our refinement network is trained on the large-scale outdoor dataset MegaDepth~\cite{Li18CVPR}, where images from 196 scenes are obtained from the Internet and then reconstructed using Structure-from-Motion~\cite{Schoenberger2016CVPR}.
We first follow the preprocessing steps from~\cite{Dusmanu2019CVPR} to regenerate camera pose labels.
We keep images with aspect ratio (width/height) between $[1.3, 1.7]$ from which we randomly select at most 500 pairs per scene which have more than 35\% visual overlap. 
Finally, we obtain in total 60661 pairs across 160 scenes. 
During training, we crop the image from the right and bottom sides so that its aspect ratio is 1.5 and then resize every image to resolution $480 \times 320$.

\PAR{Training Details.} For each training image pair, we randomly select 400 matches from the NCNet match proposals and then apply our expansion mechanism, which gives us 3200 matches to be processed by the two regressors.
The regressors are optimized using Adam~\cite{Diederik2015ICLR} with an initial learning rate of $5e^{-4}$ for 5 epochs and then $1e^{-4}$ until it converges.
Our method is implemented in Pytorch~\cite{Paszke2017NIPSW} v1.4. 
Each of our training is performed on a RTX 8000 48GB GPU.

%%%%%%%%%%%%%%%%%%%%%%%%%%

\section{Training Ablation Study.}\label{sec:supp:train_ablat}
We show {\it Patch2Pix} variants trained under different training settings including: with or without patch expansion (Exp.), different feature collection for patch pairs (Feat.), the two thresholds $\widehat{\theta}_{cls}, \widehat{\theta}_{geo}$ used to calculate the losses of the mid-level regressor, and the two $\widetilde{\theta}_{cls}, \widetilde{\theta}_{geo}$ for the fine-level regressor (\cf Sec.3.1 \& 3.2 in our main paper). 
We compare all variants when they are trained with a learning rate of $5e^{-4}$ for 5 epochs, since training longer does not change the comparison in our case.
Those variants are evaluated using HPatches~\cite{Balntas2017CVPR} for homography estimation at a confidence threshold 0.5 for ablation.
We report the percentage of correctly estimated homographies whose average corner error distance is below 1/3/5 pixels.
We give an ID to different groups of experiments for convenience and present the results in Tab.~\ref{tab:supp:exp_train_ablat}.

\PAR{Training Hyper-parameters.} As shown in Tab.~\ref{tab:supp:exp_train_ablat}, in the experiments of group I, we keep other settings identical and only modify the feature collection. We show that taking features from all layers before the last layer $f_4$ leads to the best results.
We then fix the feature collection and vary the thresholds which are used to identify the labels for classification and the subset of matches to be optimized for regression. 
Comparing models within group II, we find that models using a threshold of 400 and 50 for the mid-level regressor perform best for viewpoint changes, while the model trained with threhsold 50 is better under illumination changes and thus overall more promising.
In group III, we again fix the feature collection and apply patch expansion mechanism to our trainings and further search for the suitable thresholds. 
We find that the model trained with $\widehat{\theta}_{cls}=\widehat{\theta}_{geo}=50$ and  $\widetilde{\theta}_{cls}=\widetilde{\theta}_{geo}=5$ overall outperform others, especially under viewpoint changes, which gives the best threshold setting.

\PAR{Effect of Local Patch Expansion.}
Comparing the best models from groups II and III, we show that with the same training epochs, our best model learns faster when using patch expansion.
We further noticed that with the same thresholds, the model trained without patch expansion converges faster at a similar accuracy, while the model with patch expansion converged slower at a better accuracy.

%%%%%%%%%%%%%%%%%%%%%%%%%%%%%%%%%%%%%%%%%
\section{Experiment Details.}\label{sec:supp:exp_details}
\PAR{Homography Estimation Details}
To compute the corner correctness metric used in~\cite{Detone2018CVPRW, Sarlin2020CVPR, Wang2020ECCV}, the four corners of one image are transformed into the other image using the estimated homography to compute the distance to the four GT corners.
We report the percentage of correctly estimated homographies whose average corner error distance is below 1/3/5 pixels.
To estimate the homography from the predicted matches, we use the \textit{findHomography} function provided in pydegensac~\cite{Chum2005CVPR,Chum2003JPRS, Mishkin2015CVIU}\footnote{https://github.com/ducha-aiki/pydegensac}, which shows marginally better accuracy compared to the OpenCV~\cite{OpenCV} implementation.
We fix the RANSAC threshold as 2 pixels since it in general works better than other thresholds for all methods.
We run all methods on a GTX TITAN X 12GB GPU under our environment using their public implementations.

\PAR{Quantization Details.}
As we mentioned in the main paper, we apply quantization to our matches to evaluate on Aachen Day-Night Benchmark (v1.0)~\cite{Sattler2018CVPR,Sattler2012BMVC}.
The introduced localization pipelines, \eg HLOC~\cite{Sarlin2019CVPR}, first reconstruct a 3D model using the local features and matches and then register the queries to the built 3D model.
Therefore, such pipelines require methods to produce keypoints that are co-occurring in several retrieval pairs to work properly in the triangulation step for reconstruction.
However, our method directly regresses matches from a pair of images. 
As such, the pixel positions found in image A for a pair (A, B) might differ slightly to those found for a pair (A, C). 
% Our method aims to identify the best-matching pixel positions for a given image pair (A, B). 
% and thus our matches are not repeatable across pairs. 
In contrast, all methods that perform separate feature extraction per image will automatically have the same detections in image A for both pairs. %and matching have an advantage because the extracted keypoints are involved in matching several pairs.
Thus, they can easily be used for triangulation. 
% As a result, those matched keypoints are repeatable, which provides more consensus for the triangulation process.
To make our matches work in this setting, we quantize our matches by representing keypoints that are closer than 4 pixels to each other with their mean location, meaning we sacrifice pixel-level accuracy here.
After quantization, we remove the duplicated matches by keeping only the one with the highest confidence score.
While it is not a perfect solution, we leave it as our future work to either add a loss that enforces detecting the same positions in A for pairs (A, B) and (A, C) %where we either learn repeatable matches
or to design a localization pipeline tailored to our matches.

\PAR{InLoc Evaluation Details.}
We follow SuperPoint~\cite{Detone2018CVPRW}+SuperGlue~\cite{Sarlin2020CVPR} to evaluate on the same top-40  retrieval pairs, where they perform a temporal consistency check to restrict the retrieval, and adopt their RANSAC threshold of 48 pixels for pose estimation.
For all methods, we test their performance two different image sizes, \ie, we resize every image to have a larger side of either 1024 or 1600 pixels.
Only for SparseNCNet~\cite{Rocco2020ECCV} we also consider its default image size, \ie, 3200 pixels.
By default, the local feature detection and description methods use the nearest neighbor mutual matcher to detect matches. 
We further test those methods when using a threshold of 0.75 on their matching scores, computed by their normalized descriptors, for outlier rejection.
We present the complete results of the methods under different settings in Tab.~\ref{tab:supp:exp_inloc_full}.
For clarity, we mark their best entries \hblue{blue} which have been presented in Tab. 2 of our main paper.

With the results of the complete table, we show that SuperPoint~\cite{Detone2018CVPRW}, SuperPoint + CAPS~\cite{Wang2020ECCV} and SIFT + CAPS~\cite{Wang2020ECCV} benefit from using a threshold of 0.75 for the outlier filtering while D2Net~\cite{Dusmanu2019CVPR} and R2D2~\cite{Revaud2019NIPS} perform better without such thresholding.
In addition, we observe that SuperPoint, D2Net , SuperPoint + CAPS and Patch2Pix benefit from using a smaller image size of 1024, while SparseNCNet performs best at size of 1600 pixels.

\begin{table}[t]
\centering
\setlength{\tabcolsep}{4pt}
%\scriptsize{
\resizebox{8.5cm}{!} {
\begin{tabular}{l c c c c}
    \toprule
    \multirow{2}{*}{Method}& \multirow{2}{*}{Imsize} & \multirow{2}{*}{Supervision}& \multicolumn{2}{c}{Localized Queries (\%, 0.25$m$/0.5$m$/1.0$m$, 10$^\circ$)}\\
    &  & & DUC1 & DUC2\\
	\midrule    % 
    \hblue{SuperPoint~\cite{Detone2018CVPRW} + NN 0.75} 	& 1024 & Full &  40.4 / 58.1 / 69.7  & 42.0 / 58.8 / 69.5  \\	
    SuperPoint + NN 0.0  									& 1024 & Full & 29.8 / 48.5 / 61.6   &  32.1 / 46.6 / 56.5  \\
    SuperPoint + NN 0.75 									& 1600 & Full & 43.9 / 67.7 / 76.3   &  39.7 / 58.0 / 71.0  \\     
    
    \hblue{D2Net~\cite{Dusmanu2019CVPR} + NN 0.0 }   		& 1024  & Full & 38.4 / 56.1 / 71.2 &  37.4 / 55.0 / 64.9\\    
    D2Net + NN  0.75   										& 1024 & Full & 31.8 / 49.0 / 55.1  &  20.6 / 34.4 / 44.3 \\
    D2Net + NN  0.0   										& 1600 & Full & 34.8 / 54.5 / 68.7  &  34.4 / 50.4 / 62.6 \\ 

    \hblue{R2D2~\cite{Revaud2019NIPS} + NN 0.0} 			& 1600 & Full & 36.4 / 57.6 / 74.2 & 45.0 / 60.3 / 67.9  \\
    R2D2 + NN 0.75   										& 1600  & Full & 35.4 / 60.6 / 75.8   &  42.7 / 57.3 / 65.6\\
            
    \hblue{SuperPoint + SuperGlue~\cite{Sarlin2020CVPR}} 	& 1600 & Full & 49.0 / \textbf{68.7} / 80.8 & 53.4 / 77.1 / \textbf{82.4} \\ 
    
    \hblue{SuperPoint + CAPS~\cite{Wang2020ECCV} + NN 0.75} & 1024 & Mix & 40.9 / 60.6 / 72.7 & 43.5 / 58.8 / 68.7 \\    
    SuperPoint + CAPS + NN 0.0 								& 1024 & Mix & 39.4 / 61.6 / 72.7 & 35.1 / 50.4 / 64.1 \\ 
    SuperPoint + CAPS + NN 0.75 							& 1600 & Mix & 43.9 / 67.7 / 76.3 & 39.7 / 58.0 / 71.0 \\     
     
    \midrule   
    \hblue{SIFT + CAPS~\cite{Wang2020ECCV} + NN 0.75}  		& 1600 & Weak & 38.4 / 56.6 / 70.7 & 35.1 / 48.9 / 58.8\\
    SIFT + CAPS + NN 0.0									& 1600 & Weak & 37.9 / 56.1 / 66.7 & 30.5 / 43.5 / 53.4\\ 
    SIFT + CAPS + NN 0.75									& 1024 & Weak & 38.4 / 53.5 / 69.7 & 33.6 / 45.0 / 55.0\\     
    
    \hblue{SparseNCNet~\cite{Rocco2020ECCV} (top2k)} 			& 1600 & Weak & 41.9 / 62.1 / 72.7 & 35.1 / 48.1 / 55.0 \\
	SparseNCNet (top2k) 									& 1024 & Weak & 37.9 / 54.0 / 70.2 & 32.8 / 45.8 / 53.4 \\
    SparseNCNet (top2k) 									& 3200 & Weak & 35.4 / 50.5 / 62.1 & 24.4 / 31.3 / 35.9 \\
    \hblue{Patch2Pix (c=0.25) }        						& 1024  & Weak & 44.4 / 66.7 / 78.3  & 49.6 / 64.9 / 72.5 \\
    Patch2Pix (c=0.25)										& 1600 & Weak & 44.9 / 67.2 / 75.8  & 43.5 / 59.5 / 69.5\\
    \midrule 
	\hblue{Patch2Pix (w.SuperPoint+CAPS)} 					& 1024 & Mix & 42.4 / 62.6 / 76.3	& 43.5 / 61.1 / 71.0 \\    
    \hblue{Patch2Pix (w.SuperGlue) }   						& 1600  & Mix & \textbf{50.0} / 68.2 / \textbf{81.8} &  \textbf{57.3 / 77.9 }/ 80.2 \\    
   	\bottomrule	

\end{tabular}
}
 \vspace{3pt}
\caption{\textbf{Complete InLoc~\cite{Taira2018CVPR} Benchmark Results.} We report the percentage of correctly localized queries under specific error thresholds. Methods are evaluated inside HLOC~\cite{Sarlin2019CVPR} pipeline to share the same retrieval pairs, RANSAC threshold. We mark the best results in \textbf{bold}. For each method, we mark its best entry among all settings in \hblue{blue} which corresponds to its result presented in Tab. 2 of our main paper.}
\label{tab:supp:exp_inloc_full}
\end{table}

\section{Qualitative Results.}\label{sec:supp:qualitative}
\begin{figure*}[t]
\centering
  \includegraphics[width=0.85\textwidth]{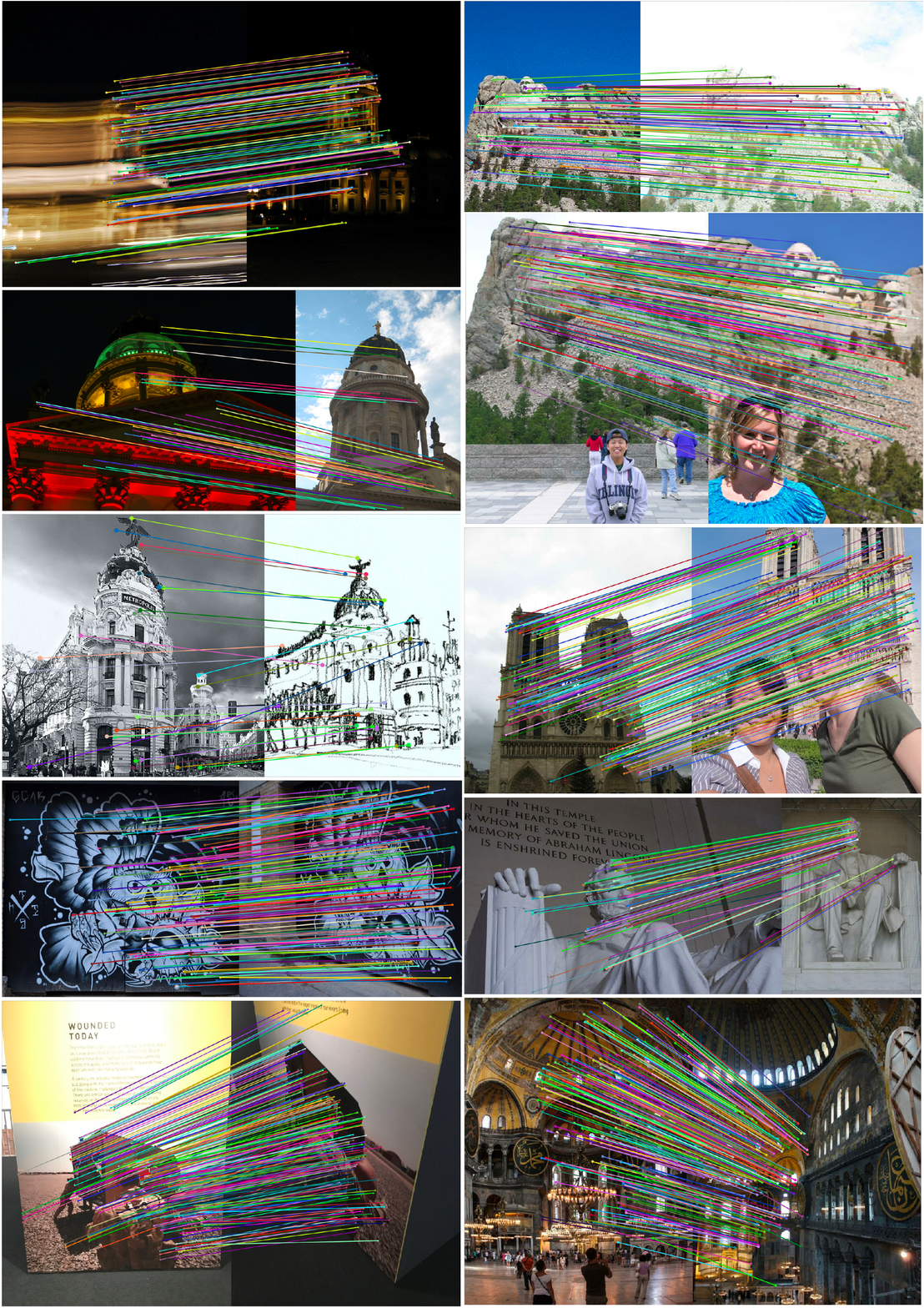}
  \caption{Example matches of \textit{Patch2Pix} on the image pairs obtained from the internet, HPatches~\cite{Balntas2017CVPR} and PhotoTourism~\cite{Jin2020IJCV}. \textit{Patch2Pix} can robustly handle strong illumination changes, large viewpoint variations, and repetitive structures.}  
  \label{fig:samples_nc_mixed}
%   \vspace{-9pt}
\end{figure*}

In \figref{fig:samples_nc_mixed}, we plot the matches estimated by \textit{Patch2Pix} on the image pairs obtained from the internet, HPatches~\cite{Balntas2017CVPR} and PhotoTourism~\cite{Jin2020IJCV}.
We use the default setting of our model, \ie, NCNet proposals and confidence score 0.25, to predict matches from the image pairs.
We identify the inlier matches using the \textit{findHomography} or \textit{findFundamentalMatrix} function provided in pydegensac~\cite{Chum2005CVPR,Chum2003JPRS, Mishkin2015CVIU}.
For the HPatches image pairs, we use \textit{findHomography} with a ransac threshold of 2.
For other image pairs, we use \textit{findFundamentalMatrix} with a RANSAC threshold of 1.
Finally, we plot at most 300 matches for each pair for clear visualization.

\begin{figure*}[t]
\centering
  \includegraphics[width=0.85\textwidth]{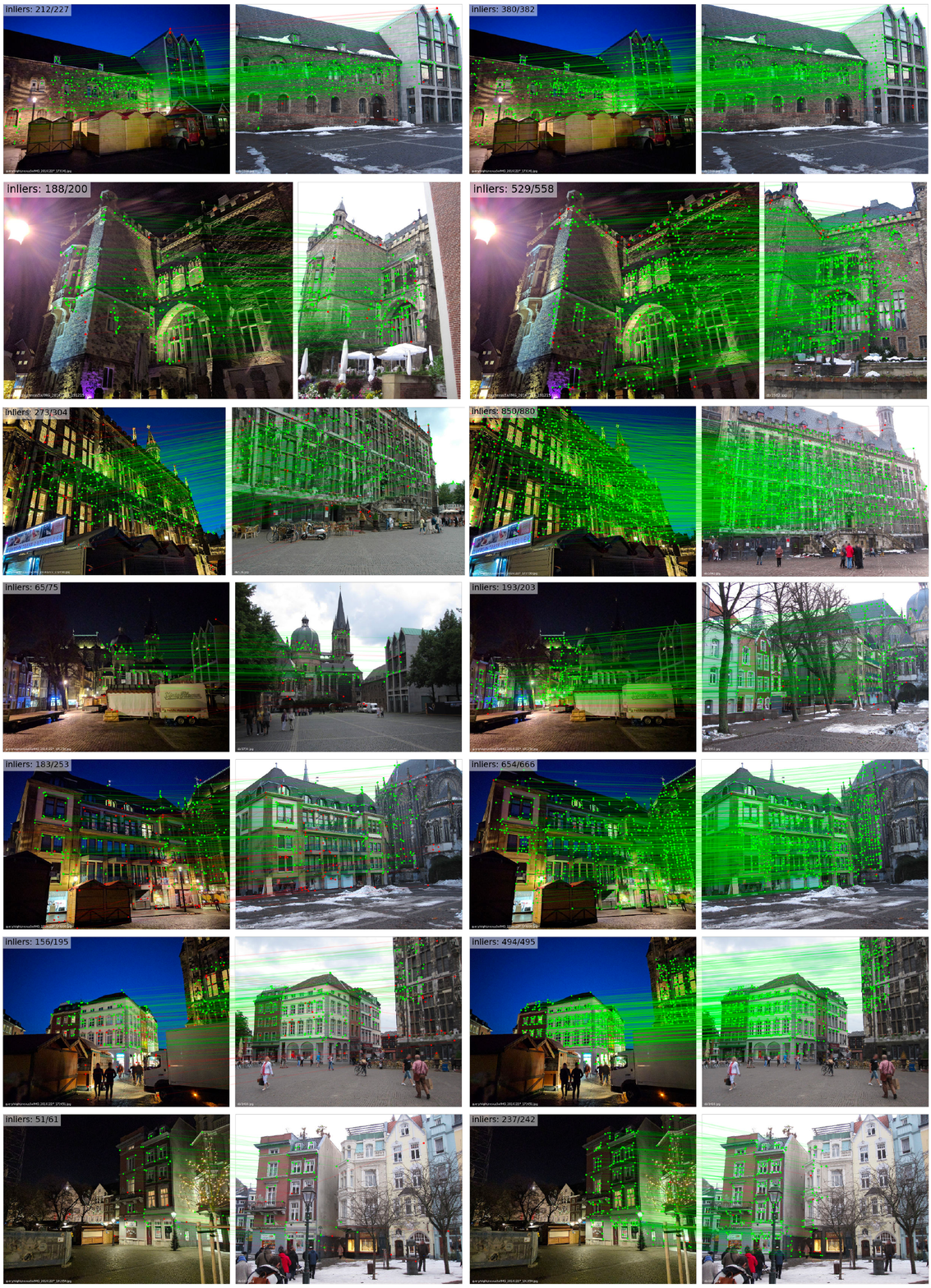}
  \caption{Example matches of \textit{Patch2Pix} using NCNet proposals (left) and SuperPoint~\cite{Detone2018CVPRW} + SuperGlue~\cite{Sarlin2020CVPR} (right) proposals on night queries of Aachen Day-Night(v1.0)~\cite{Sattler2018CVPR,Sattler2012BMVC}.}  
  \label{fig:samples_nc_sg_aachen}
%   \vspace{-9pt}
\end{figure*}

\begin{figure*}[t]
\centering
  \includegraphics[width=0.85\textwidth]{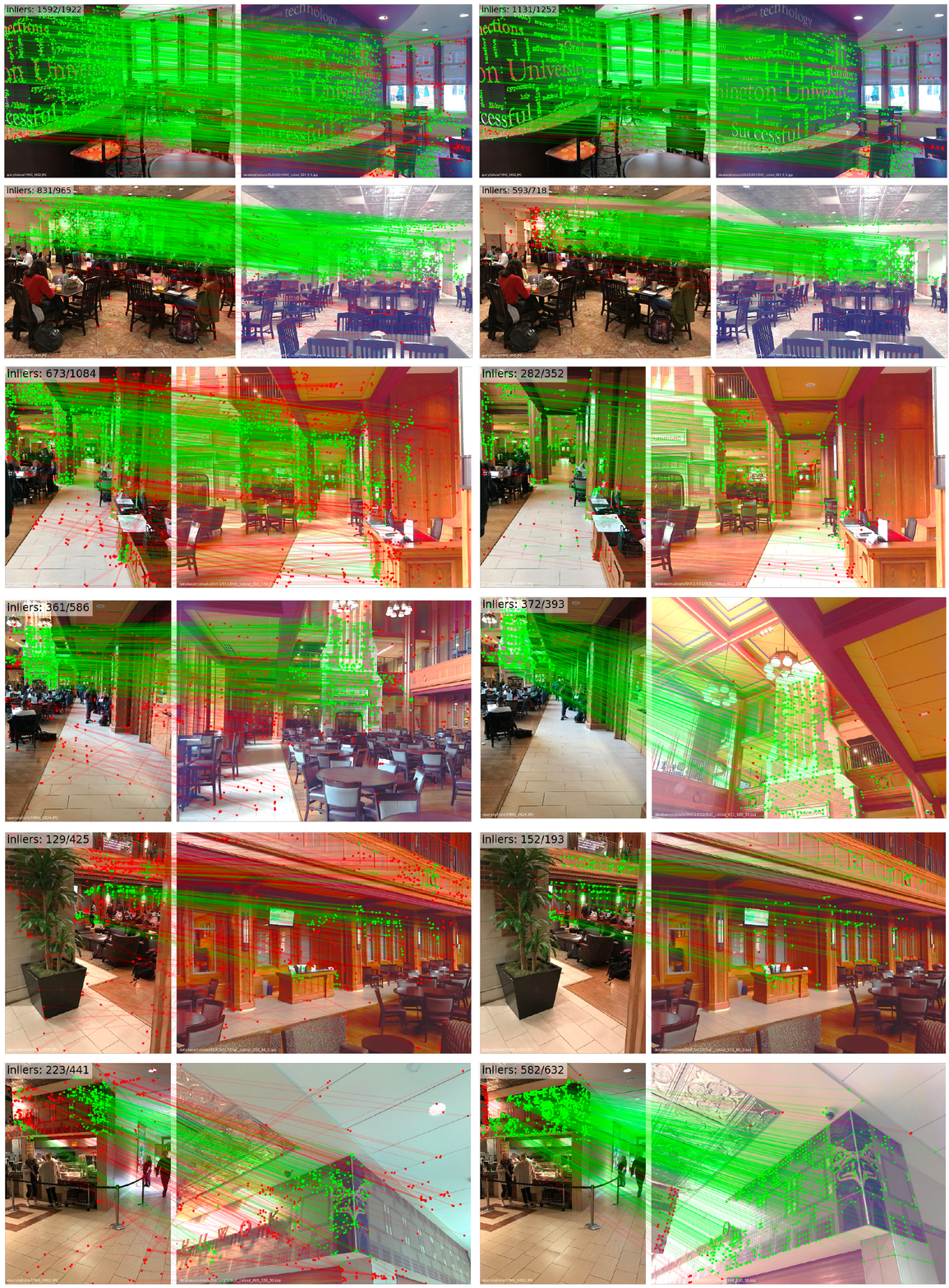}
  \caption{Example matches of \textit{Patch2Pix} using NCNet proposals (left) and SuperPoint~\cite{Detone2018CVPRW} + SuperGlue~\cite{Sarlin2020CVPR} (right) proposals on InLoc~\cite{Taira2018CVPR} DUC1.}  
  \label{fig:samples_nc_sg_inloc_duc1}
%   \vspace{-9pt}
\end{figure*}

\begin{figure*}[t]
\centering
  \includegraphics[width=0.85\textwidth]{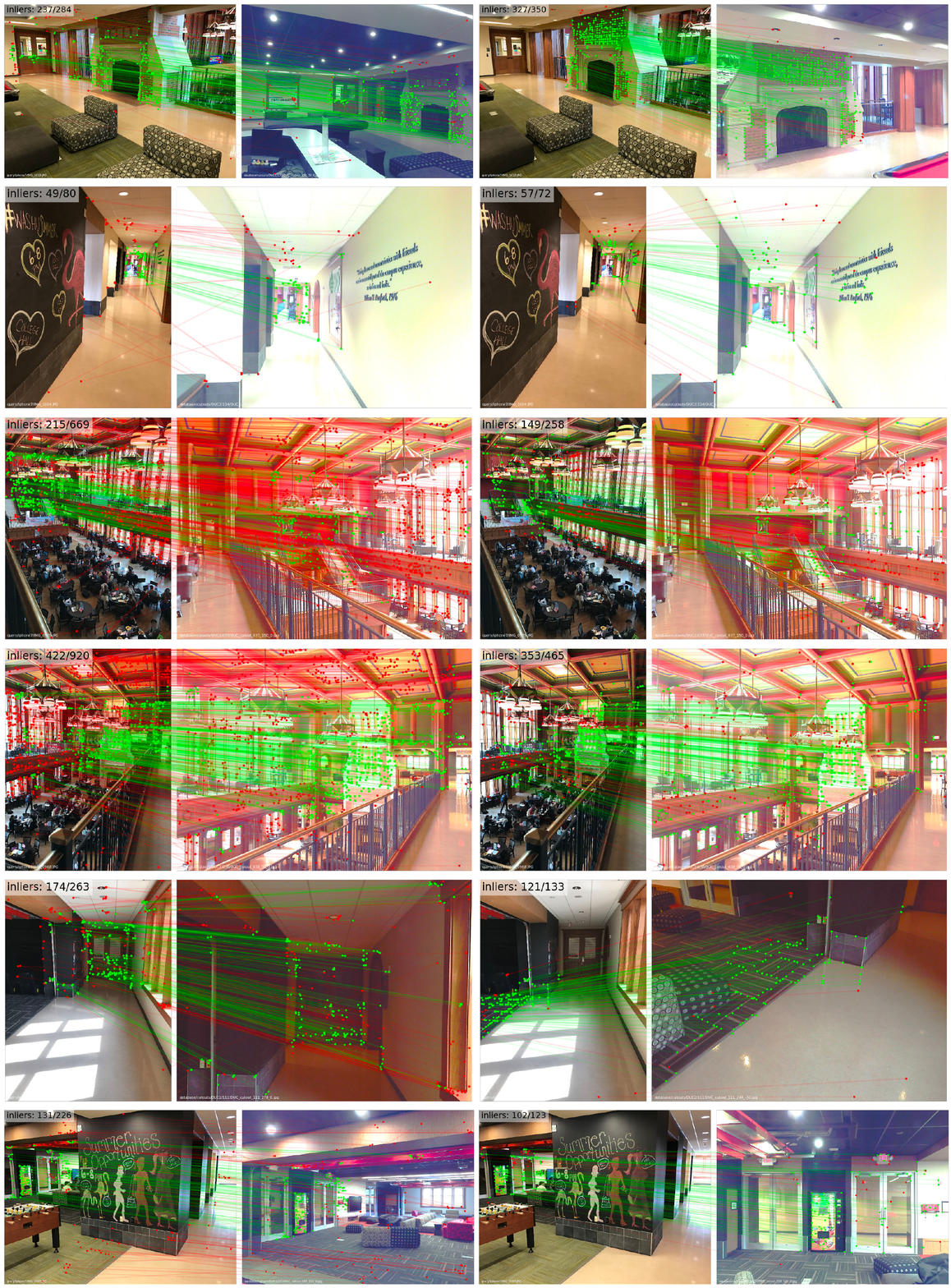}
  \caption{Example matches of \textit{Patch2Pix} using NCNet proposals (left) and SuperPoint~\cite{Detone2018CVPRW} + SuperGlue~\cite{Sarlin2020CVPR} (right) proposals on InLoc~\cite{Taira2018CVPR} DUC2.}  
  \label{fig:samples_nc_sg_inloc_duc2}
%   \vspace{-9pt}
\end{figure*}

Furthermore, we visualize the matches on Aachen Day-Night (v1.0)~\cite{Sattler2018CVPR,Sattler2012BMVC} in \figref{fig:samples_nc_sg_aachen} and on InLoc~\cite{Taira2018CVPR} in \figref{fig:samples_nc_sg_inloc_duc1} and  \figref{fig:samples_nc_sg_inloc_duc2}.
We show the matches refined by \textit{Patch2Pix} when we use our NCNet baseline, and when we use SuperPoint~\cite{Detone2018CVPRW} + SuperGlue~\cite{Sarlin2020CVPR} for the match proposals.
For a randomly selected query, we pick the database images with the most inlier matches identified by the camera pose solver during localization.
We plot the inliers in \textcolor{green}{green} and other matches in \textcolor{red}{red} and count the inlier numbers.

% The qualitative results show that our method performs well in both indoor and outdoor scenes and robustly handles strong illumination changes, large viewpoint variations, and repetitive structures.

{\small
\bibliographystyle{ieee_fullname}
\bibliography{localize-cvpr21.bib}
}

\end{document}